\documentclass{article}

\usepackage{microtype}
\usepackage{graphicx}
\usepackage{subfigure}
\usepackage{booktabs} 
\usepackage{pifont}
\usepackage{colortbl}
\usepackage{adjustbox}
\usepackage{multicol}
\usepackage{multirow}
\usepackage{ulem}
\usepackage{capt-of}
\usepackage{rotating}
\usepackage{tabularx}
\usepackage{booktabs}

\usepackage{hyperref}

\newcommand{\method}[0]{{Pruner-Zero}}

\newcommand{\cmark}{\ding{51}}
\newcommand{\xmark}{\ding{55}}

\usepackage[accepted]{icml2024}

\usepackage{amsmath}
\usepackage{amssymb}
\usepackage{mathtools}
\usepackage{amsthm}
\usepackage{colortbl}

\usepackage{xcolor}
\usepackage[capitalize,noabbrev]{cleveref}

\definecolor{Gray}{gray}{0.95}

\newcommand{\gr}{\rowcolor[gray]{.95}}

\newcommand{\wc}{\cellcolor{white}}

\newcommand{\tbf}{\textbf}

\theoremstyle{plain}

\theoremstyle{definition}

\theoremstyle{remark}

\icmltitlerunning{\method: Evolving Symbolic Pruning Metric From Scratch for Large Language Models}

\begin{document}

\twocolumn[
\icmltitle{\method: Evolving Symbolic Pruning Metric from scratch for Large Language Models}

\icmlsetsymbol{equal}{*}

\begin{icmlauthorlist}
\icmlauthor{Peijie Dong}{hkustgz,equal}
\icmlauthor{Lujun Li}{hkust,equal}
\icmlauthor{Zhenheng Tang}{hkbu,hkustgz}
\icmlauthor{Xiang Liu}{hkustgz}
\icmlauthor{Xinglin Pan}{hkustgz}
\icmlauthor{Qiang Wang}{hit}
\icmlauthor{Xiaowen Chu}{hkustgz,hkust}
\end{icmlauthorlist}

\icmlaffiliation{hkustgz}{The Hong Kong University of Science and Technology (Guangzhou)}
\icmlaffiliation{hkust}{The Hong Kong University of Science and Technology}
\icmlaffiliation{hkbu}{Hong Kong Baptist Univeristy}
\icmlaffiliation{hit}{Harbin Institude of Technology, Shenzhen}

\icmlcorrespondingauthor{Qiang Wang}{qiang.wang@hit.edu.cn}
\icmlcorrespondingauthor{Xiaowen Chu}{xwchu@ust.hk}

\icmlkeywords{Large Language Model, Model Pruning, Symbolic Regression, Model Compression, Post-training Pruning}
\vskip 0.2in
]

\printAffiliationsAndNotice{\icmlEqualContribution} 

\begin{abstract}
Despite the remarkable capabilities, Large Language Models (LLMs) face deployment challenges due to their extensive size. Pruning methods drop a subset of weights to accelerate, but many of them require retraining, which is prohibitively expensive and computationally demanding. Recently, post-training pruning approaches introduced novel metrics, enabling the pruning of LLMs without retraining. However, these metrics require the involvement of human experts and tedious trial and error. To efficiently identify superior pruning metrics, we develop an automatic framework for searching symbolic pruning metrics using genetic programming. In particular, we devise an elaborate search space encompassing the existing pruning metrics to discover the potential symbolic pruning metric. We propose an opposing operation simplification strategy to increase the diversity of the population. In this way, \method{} allows auto-generation of symbolic pruning metrics. Based on the searched results, we explore the correlation between pruning metrics and performance after pruning and summarize some principles. Extensive experiments on LLaMA and LLaMA-2 on language modeling and zero-shot tasks demonstrate that our \method{} obtains superior performance than SOTA post-training pruning methods. Code at: \url{https://github.com/pprp/Pruner-Zero}.
\end{abstract}

\section{Introduction}

\begin{table*}[t]
    \centering
    \caption{The existing pruning metrics tailored for LLMs. ``W" denotes weight update, and ``C" denotes the calibration data. ``UOP" denotes unary operations, and ``BOP" denotes binary operations. $\sigma$ denotes the min-max scaling operation.}\label{tab:property}
    \renewcommand{\arraystretch}{1.02}
    \setlength{\tabcolsep}{4.5pt}
    \resizebox{.85\textwidth}{!}{
    \begin{tabular}{l c c c c c}
            & & &  & \\
        \toprule
        \textbf{Method}  & \textbf{W} & \textbf{C} &  \textbf{Pruning Metric} $\mathcal{S}$  & \textbf{UOP} & \textbf{BOP} \\
        \midrule
        Magnitude~\cite{han2015deep_compression_magnitude}  & \xmark & \xmark & $|\mathbf{W}|$ & $|\cdot|$ & $\emptyset$ \\
        SparseGPT~\cite{Frantar2023SparseGPTML} & \cmark & \cmark & $\left[|\mathbf{W}|^{2} / \mathrm{diag}\bigl[\mathbf{H}^{-1}\bigr]\right]_{ij}$ & $|\cdot|$, $(\cdot)^2$ & $\div$ \\
        Wanda~\cite{Sun2023ASA_wanda}  & \xmark & \cmark & $|\mathbf{W}_{ij}| \cdot \|\mathbf{X}_{j}\|_{2}$ 
        & $|\cdot|, ||\cdot||_2$ & $\times $\\
        GBLM-Pruner$_{l1}$~\cite{Das2023BeyondSH_GBLM} & \xmark & \cmark & $|\mathbf{W}| \cdot ||\mathbf{G}||_1$ &  $|\cdot|$, $||\cdot||_1$ & $\times$ \\
        GBLM-Pruner$_{l2}$~\cite{Das2023BeyondSH_GBLM} & \xmark & \cmark & $|\mathbf{W}| \cdot ||\mathbf{G}||_2$ & $|\cdot|$, $||\cdot||_2$ & $\times$ \\
        \gr \method{} & \xmark & \cmark & $\left|\left|\mathbf{W}\right|\times \left|\mathbf{W}\right|\right| \times \sigma(|\mathbf{G}|)$ & $|\cdot|, \sigma(\cdot)$ & $\times$ \\
        \bottomrule
    \end{tabular}}
    \vskip -0.15in
\end{table*}

Recent breakthroughs in Large Language Models (LLMs)~\cite{openai2023gpt4, touvron2023llama, touvron2023llama2, Abdin2024Phi3TR} have revolutionized the field of Natural Language Processing~(NLP) tasks, enabling remarkable progress across a wide spectrum of tasks, including both Natural Language Understanding and Natural Language Generation. A key differentiator of LLMs from prior convolution-based models is their unprecedented scale. With billions of parameters, LLMs can effectively manage complex tasks and have exhibited exceptional performance. However, the vast number of parameters that these models possess necessitates substantial computational resources, highlighting a critical challenge in their deployment. For instance, deploying models like GPT-3, with its 175 billion parameters, underscores the intense computational demands, necessitating advanced GPU technologies. 

Several model compression techniques are developed to solve these problems, such as model quantization~\cite{Frantar2022GPTQAP, liu2023llm_qat, NEURIPS2022_adf7fa39_zero_quant}, model sparsing~\cite{Frantar2023SparseGPTML, Sun2023ASA_wanda}, and knowledge distillation~\cite{li2022norm,lishadow,li2022self}. Model sparsing compresses the model by identifying and eliminating redundant elements in the weight matrix. Model sparsing is one of the most promising techniques for model deployment due to its flexibility. 
Previous research works on model sparsing require training from random initialization~\cite{hoang2023revisiting, sreenivasan2022rare,louizos2018learning}, retraining process~\cite{liu2018rethinking, chen2023otov} or extensive iterative pruning~\cite{NEURIPS2021_23e582ad, tanaka2020pruning}. However, the inherent complexities, along with the substantial computational and data requirements of LLMs, present significant challenges that render these conventional sparse strategies less feasible.

Given the extensive data corpus and substantial model dimensions required by LLMs, post-training pruning has become an increasingly crucial methodology. Due to its minimal resource demands, this approach is highly advantageous, offering a cost-efficient alternative for optimizing LLMs. The development of the post-training pruning method, as highlighted in recent studies~\cite{miao2022learning, Frantar2023SparseGPTML, Sun2023ASA_wanda}, marks a significant advancement in this field. These methods streamline the pruning process, further reducing the resource requirements and making LLMs more democratized and accessible. SparseGPT~\cite{Frantar2023SparseGPTML} is proposed to conduct post-training pruning for LLMs, which allows for a significant reduction in the size of these models to prune the GPT family to 50\% sparsity. Wanda~\cite{Sun2023ASA_wanda} does not require retraining or weight updates, which makes the pruned LLM immediately ready for inference tasks. GBLM-Pruner~\cite{Das2023BeyondSH_GBLM} further dives into the design of the pruning metric by emphasizing the importance of first-order information, a.k.a. gradient. As shown in Table~\ref{tab:gblm-pruner-table}, we identify two key challenges of the existing pruning metrics: 
(1) \textbf{Human-Dependence}: existing methods heavily rely on domain knowledge and thus require massive efforts of trial and error. (2) \textbf{Format-Sensitivity}: As illustrated in Table~\ref{tab:gblm-pruner-table}, pruning metrics demonstrate a pronounced sensitivity to their format, demanding a rigorous and systematic approach to experimentation. These points of contention raise pivotal inquiries:

\textit{(1) How to \textbf{formulate and devise comprehensive pruning metrics} that encapsulate the strengths of existing ones? }

\textit{(2) How to find the \textbf{optimal pruning metric} tailored for Large Language Models?}

To address the \textbf{first question}, we present \textbf{a comprehensive search space} through an exhaustive review of the existing pruning metrics, as detailed in Table~\ref{tab:property}. We meticulously dissect the structure of these metrics by analyzing their constituent inputs and operations. The inputs considered are Weight~(W), Gradient~(G), Hessian~(H), and Activation~(X), while the operations are categorized into unary or binary ones. Given the computational complexity of the Hessian metric, which scales quadratically with the dimensionality of the hidden layer, $(O(d^2_{hidden}))$, we opt to exclude it from our considerations within this paper. For symbolic operations, we collect the operation utilized in the existing pruning metrics to build the operation vocabulary in Table~\ref{appendix:operation_vocabulary}.
Inspired by Symbolic Regression (SR), we observe that the pruning metrics are composed of a series of symbols, and thus can be represented as an expression tree, here we call it a symbolic pruning metric. 
We formulate these pruning metrics as \textbf{Symbolic Pruning Metrics (SPM)}, denoting them as a combination of symbolics. However, SR is a complex combinatorial problem, as the search space for equations grows exponentially with the number of its operations, which comes to the second question. 

\begin{figure*}
\centering
\includegraphics[width=0.9\linewidth]{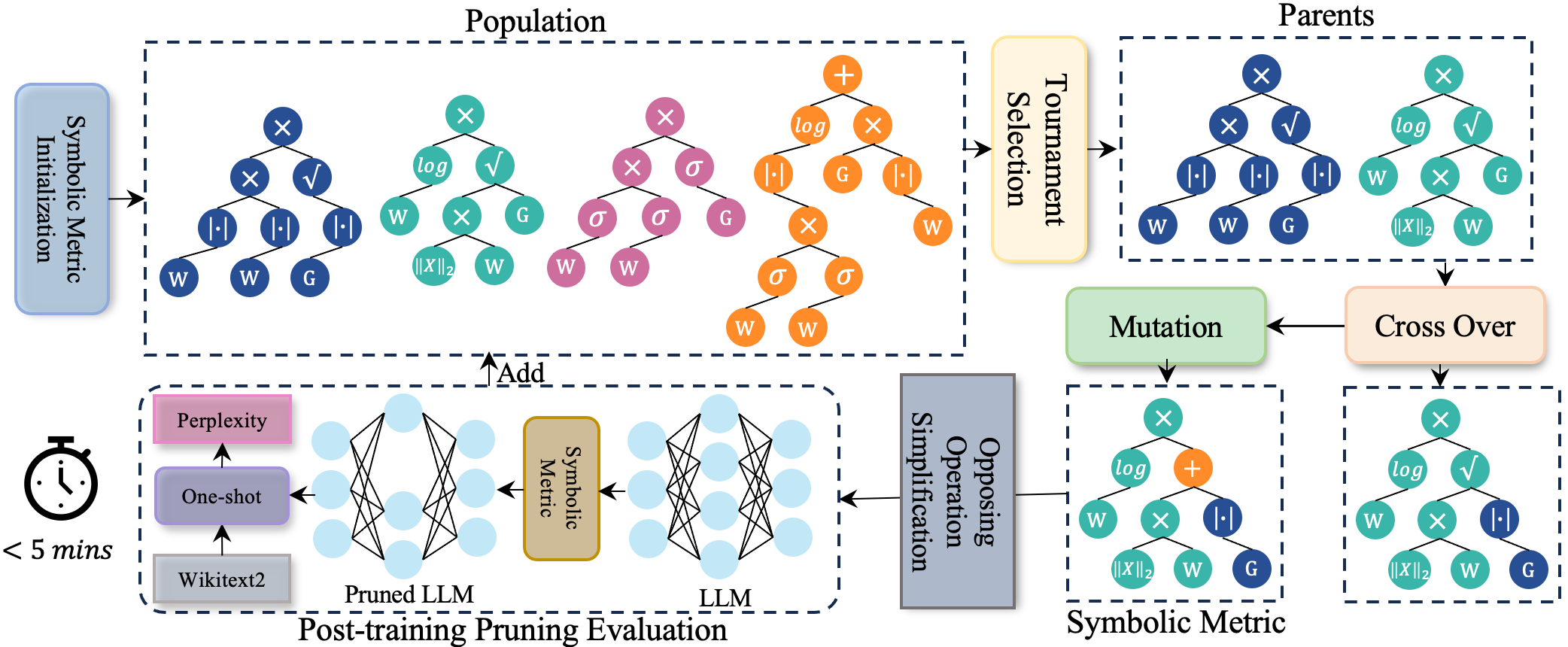}
    \label{fig:main_figure}
\vskip -0.15in
\caption{Overview of the Automatic Symbolic Pruning Metric Discovery Process in our \method{} framework. This process employs genetic programming to iteratively generate and refine symbolic pruning metrics via tournament selection, subtree crossover, and node mutation. Upon generating offspring, the Opposing Operation Simplification (OOS) strategy is applied to diminish repetition. Subsequently, evaluation is conducted on the LLaMA-2-7B using the WikiText2 dataset, with perplexity serving as the fitness metric. Note that it only takes less than 5 minutes to perform the post-training pruning evaluation.}
\vspace{-0.3cm}
\end{figure*}

In addressing the \textbf{second question}, we introduce \textbf{the \method{} framework}, which employs Genetic Programming (GP) to devise a symbolic pruning metric (SPM) encapsulated within a tree structure, as depicted in Figure~\ref{fig:main_figure}. The terminal nodes represent variables such as Weight (W), Gradient (G), and Activation (X), while its internal nodes represent mathematical operations. This innovative method facilitates the organic growth of an expression tree that emulates the principles of biological evolution, complete with node mutations and subtree crossovers that effectively probe the diversity of potential pruning metrics. We conduct fast post-training pruning evaluations for each SPM for LLaMA-7B on the WikiText2 dataset to get the perplexity of its fitness in less than 5 minutes. 
However, we recognize the existence of correlations among certain operations, particularly those that are inversely related, which complicates the search space with multiple symbolic trees that, despite their differences, are mathematically equivalent. To counteract this challenge, we propose the Opposing Operation Simplification (OOS) strategy. This strategy is designed to pinpoint and streamline opposing patterns, thereby curbing the redundancy within the search space and enhancing the efficiency of the metric discovery process. 

Upon identifying the most promising SPM candidate, we adapt it across a wider spectrum of LLM families, including LLaMA~\cite{touvron2023llama}, LLaMA-2~\cite{touvron2023llama2}, Tiny-LLaMA~\cite{Zhang2024TinyLlamaAO}, and OPT~\cite{Zhang2022OPTOP}. Utilizing this framework, we delve deeper into the relationship between low perplexity and the design of pruning metrics. This exploration involves a comprehensive analysis of the correlations among different operations. From this analysis, we can distill and summarize key principles for developing effective pruning metrics. These principles are in harmony with the insights gained from our previously identified SPM, providing a cohesive and informed approach to pruning metric design. Our contributions are summarized as follows:

\noindent 1. We formulate the pruning metric discovery as a Symbolic Regression problem and propose a unified and comprehensive search space that encompasses existing pruning metrics (Sec.~\ref{sec:search_space_design}).

\noindent 2. We introduce \method{}, the first Symbolic Pruning Metric search framework for the automated discovery of optimal pruning metrics tailored for LLMs utilizing Genetic Programming. We propose the Opposing Operation Simplification (OOS) strategy to reduce the redundancy in the search space (Sec.~\ref{genetic_programming_framework}).

\noindent 3. Our comprehensive experiments on language modeling and zero-shot tasks demonstrate that our \method{} can achieve SOTA performance without retraining or weight update for LLMs (Sec.~\ref{experiments}). 

\noindent 4. We further investigate the correlation between the performance and the SPM design and uncover some principles in designing pruning metrics (Sec.~\ref{sec:analysis}).

\section{Related Work}
\subsection{Language Model Pruning} 
Network pruning can significantly reduce model complexity while maintaining performance, albeit typically necessitating extensive retraining. Given the substantial parameter sizes and extensive datasets of LLMs, traditional pruning methods~\cite{hoang2023revisiting, sreenivasan2022rare, liu2018rethinking, chen2023otov,NEURIPS2021_23e582ad} become impractical. In this paper, we focus on the post-training pruning techniques.

\textbf{Problem Statement.} Post-training pruning serves as a practical scenario where we are given a well-optimized $W$ and a symbolic pruning metric $\mathcal{S}$. Post-training compression is conducted by splitting the full-model compression problem into layer-wise sub-problems. 
Specifically, for post-training pruning, our objective is to find a sparsity mask (binary mask) $M_l$ for each layer $l$ with a certain target sparsity ratio to minimize the $l2$-error, formulated as:
\begin{equation}
    \textbf{argmin}_{M_{l}}||W_lX_l-(M_l \odot W_l) X_l||^2_2
\end{equation}
Here, the mask $M_l$ is determined by symbolic pruning metric $\mathcal{S}$ through $M_l=f(\mathcal{S}, W_l, \phi)$. The function $f$ ranks the weights in $W_l$ according to their importance as per $\mathcal{S}(W_l, X_l, G_l)$, and then selects the most significant weights up to the sparsity ratio $\phi$. For example, a basic pruning metric is magnitude~\cite{song2016deep_compression}, where $\mathcal{S}(W_l)=|W_l|$ employs the element-wise absolute value to assess weight significance. Thus, the key to post-training pruning lies in the symbolic pruning metric. 

\textbf{Post-training Pruning for LLMs.} Deep Compression~\cite{han2015deep_compression_magnitude} proposed magnitude-based pruning, which eliminates weights with the smallest absolute values, based on their minimal impact on network output. SparseGPT~\cite{Frantar2023SparseGPTML} proposed a framework for post-training pruning that obviates the need for retraining by using Hessian matrices and calibration data to update weights. Complementing this, Wanda~\cite{Sun2023ASA_wanda} streamlines the process by simplifying SparseGPT's methodology. Additionally, GBLM-Pruner~\cite{Das2023BeyondSH_GBLM} employs the first-order term of the Taylor expansion, emphasizing the significance of gradients. Structured pruning methods, such as activation pruning and neuron/filter output statistics, are used for GPU acceleration. LLM-Pruner~\cite{Ma2023LLMPrunerOT} examines model dependencies, incorporating both first-order and approximated Hessian information. LLM Surgeon~\cite{Ouderaa2023TheLS} adapts Kronecker-factored curvature approximations to LLMs, targeting 20\%-25\% low sparsity. In this paper, our focus is on the post-training pruning of language models without retraining or weight updates.

\textbf{Efficient and Low Resource Compression.} 
Due to the large size of language models, there is an increasing demand for efficient LLM compression without using the original training data. As for efficient compression,~\cite{kwon2022a} accelerates the post-training by defining the reconstruction error as a linear least squares problem. SparseGPT~\cite{Frantar2023SparseGPTML} and GPTQ~\cite{Frantar2022GPTQAP} propose the layer-wise optimal brain surgeon. Due to the constraint of availability of the training corpus, data-free methods~\cite{Srinivas2015DatafreePP, Yvinec2021REDD} prune the neural network by measuring the similarity of neurons. Most related to our approach is pruning with limited data, which requires no modification to the original training procedure and no retraining of the pruned network on the full training datasets. To mitigate the accuracy drop, a layer-wise reconstruction problem is involved to minimize the change of output evaluated on the calibration data.

\subsection{Symbolic Regression} 

Symbolic Regression (SR) is a distinctive approach within genetic programming, aiming at discovering the analytical expression that accurately models a complex dataset without the need for pre-specified functional forms. This adaptive nature of SR allows it to dynamically evolve when fitting data, making it especially useful in scenarios where the underlying relationships are complex or unknown. Originating from genetic programming~\cite{Koza1994GeneticPA}, which applies principles of natural selection to evolve computer programs, SR has evolved into an essential tool for modeling and prediction. It stands out for its flexibility in handling complex modeling tasks, surpassing traditional regression methods in adaptability and interpretability.

The integration of deep learning techniques with SR, particularly through approaches like Equation Learner (EQL)~\cite{Martius2016ExtrapolationAL, Sahoo2018LearningEF, Werner2021InformedEL}, marks a significant advancement in the field. EQL and similar methods employ neural networks, utilizing basic mathematical operations as activation functions to construct and train networks until a sparse and efficient structure is achieved. The resulting structure is then unrolled into a final symbolic expression. This synthesis of neural network capabilities with symbolic regression illustrates the potential for scalable and flexible model generation, enhancing the applicability of SR in various complex scenarios. Notably, in this paper, we pioneer the conceptualization of the pruning metric problem within the framework of symbolic regression.

\section{\method{} Framework} 

In this section, we first introduce the search space design of pruning metrics and then present the details of the \method{} framework to search for symbolic pruning metrics.

\begin{algorithm}[t]
\caption{Evolution Search for \method{}}
\label{alg:evolution}
\textbf{Input}: Search space $\mathcal{S}$, population size $|\mathcal{P}|$, sample ratio $r$, top-k $k$, selection ratio $\rho$, max iteration $\mathcal{N}$, tree depth $d$. \\
\textbf{Output}: Best symbolic tree with lowest Perplexity (PPL) after post-training pruning evaluation.\\
\begin{algorithmic}[1]
\STATE Initialize promising candidates $\mathcal{C}$ to $\emptyset$
\STATE $\mathcal{P}_0 \gets$ \textit{SymbolicMetricInitialization}$(P_i)$
\FOR{$i = 1$ $\cdots$ $\mathcal{N}$}
    \STATE Clear promising candidates $\mathcal{C}$ to $\emptyset$
    \STATE Randomly select $r \times |\mathcal{P}|$ subnets $\hat{P}_i \in \mathcal{P}$ to get $\mathcal{C}$
    \STATE Candidates $\{\mathcal{S}_i\}_{k} \gets$ \textit{GetTopk}$(\mathcal{C}, k)$
    \STATE Parent $\mathcal{S}_i^1, \mathcal{S}_i^2 \gets$ \textit{RandomSample}$(\{\mathcal{S}_i\}_{k})$
    \STATE $\mathcal{S}_i^c \gets$ \textit{CROSSOVER}$(\mathcal{S}_i^1, \mathcal{S}_i^2)$
    \STATE $\mathcal{S}_i^m \gets$ \textit{MUTATE}$(\mathcal{S}_i^c)$ with probability $p$
    \STATE $\mathcal{S}_i^s \gets$ \textit{OOS}$(\mathcal{S}_i^m)$
    \IF{\textit{IsEquivalent}$(\mathcal{S}_i^s, \mathcal{S}_i^1)$ OR \textit{IsEquivalent}$(\mathcal{S}_i^s, \mathcal{S}_i^2)$}
        \STATE $\mathcal{S}_i^s \gets$ \textit{RandomSample}$(\mathcal{S}, d)$
    \ENDIF
    \STATE Append offspring $\mathcal{S}_i^s$ to population with its PPL
    \STATE Remove the symbolic tree with the highest PPL
\ENDFOR

\STATE \textbf{Procedure} \textit{OOS}$(\mathcal{S}_i^m)$
\FOR{each node in $\mathcal{S}_i^m$}
    \STATE Determine if the current node is opposing operations compared to $\mathcal{S}_i^m$.
    \IF{opposing operation is found}
        \STATE Remove the opposing operation pair.
    \ENDIF
\ENDFOR
\STATE \textbf{end Procedure}
\end{algorithmic}
\end{algorithm}

\subsection{\method{} Search Space Design}\label{sec:search_space_design}

To ensure the effectiveness of \method, we devise a unified framework that encompasses and extends beyond existing pruning metrics. This framework's search space includes three types of inputs and 17 primitive operations, enabling us to reconstruct existing pruning metrics as detailed in Table~\ref{tab:property} and Table~\ref{tab:gblm-pruner-table}.

\noindent \textbf{LLM Statistics as Inputs.} 
Our search space comprises three specific input types: activations (X), gradients (G), and weights (W). We exclude the Hessian matrix in SparseGPT~\cite{Frantar2023SparseGPTML} due to its quadratic complexity. Given the substantial size of LLMs, we collect and preprocess gradient information using $128$ calibration samples and archive it locally. Consequently, during the search phase, only activation and weight are required, thereby expediting the search process.

\noindent \textbf{Primitive Operations.} 
Inspired by AutoML-Zero~\cite{real2020automlzero} and EMQ~\cite{dong2023emq}, we utilize a comprehensive suite of primitive operations that include unary and binary ones. These operations are informed by existing symbolic pruning metrics, as detailed in Table~\ref{tab:property}. The complete primitive operations are available in Appendix~\ref{sec:operation_vocabulary}, with an in-depth analysis provided in Sec.~\ref{sec:analysis} and Appendix~\ref{appendix:discuss}.

\noindent \textbf{Pruning Metric as Expression Tree.} 
We define the pruning metrics, as outlined in Table~\ref{tab:property}, through a structured combination of operations and Large Language Model (LLM) statistics. In this framework, LLM statistics are situated at the leaf nodes, while primitive operations occupy the internal nodes. Binary operations are characterized by two child nodes, whereas unary operations possess a single child node, with the second node represented by a ``\#" placeholder. The root node, designated as a primitive operation, yields a singular output. Within this expression tree, the output's dimensions are required to align with those of the weights, thereby indicating the saliency of each weight.

\subsection{Genetic Programming Framework}
\label{genetic_programming_framework}

We conceptualize the identification of symbolic pruning metrics as a challenge within Symbolic Regression (SR), recognized as a complex combinatorial NP-hard problem~\cite{virgolin2022symbolic}. This complexity is attributed to the search space for pruning metrics, which expands exponentially with the increase in the number of primitive operations. Genetic Programming (GP), being the predominant methodology for addressing SR problems, guides our approach. Accordingly, we leverage GP to iteratively search for the optimal symbolic pruning metric.

An overview of our framework is depicted in Figure~\ref{fig:main_figure} and elaborated upon in Algorithm~\ref{alg:evolution}. Initially, we populate $\mathcal{P}$ with symbolic trees of varying depths. In each iteration, two parent symbolic pruning metrics are selected via tournament selection from the top-$k$ candidates $(\{\mathcal{S}_i\}_{k})$ with \textit{GetTopk($\mathcal{C}, k$)} in line 6 of Algorithm~\ref{alg:evolution}. These parents then undergo subtree crossover and node mutation at a probability $p$, generating offspring symbolic pruning metrics. To minimize population redundancy, we implement the Opposing Operation Simplification (OOS) strategy, ensuring uniqueness among the metrics within the population.

\begin{figure}[t]
\vskip 0.15in
    \centering
    \begin{minipage}[b]{0.49\linewidth}
        \centering
        \includegraphics[width=\linewidth]{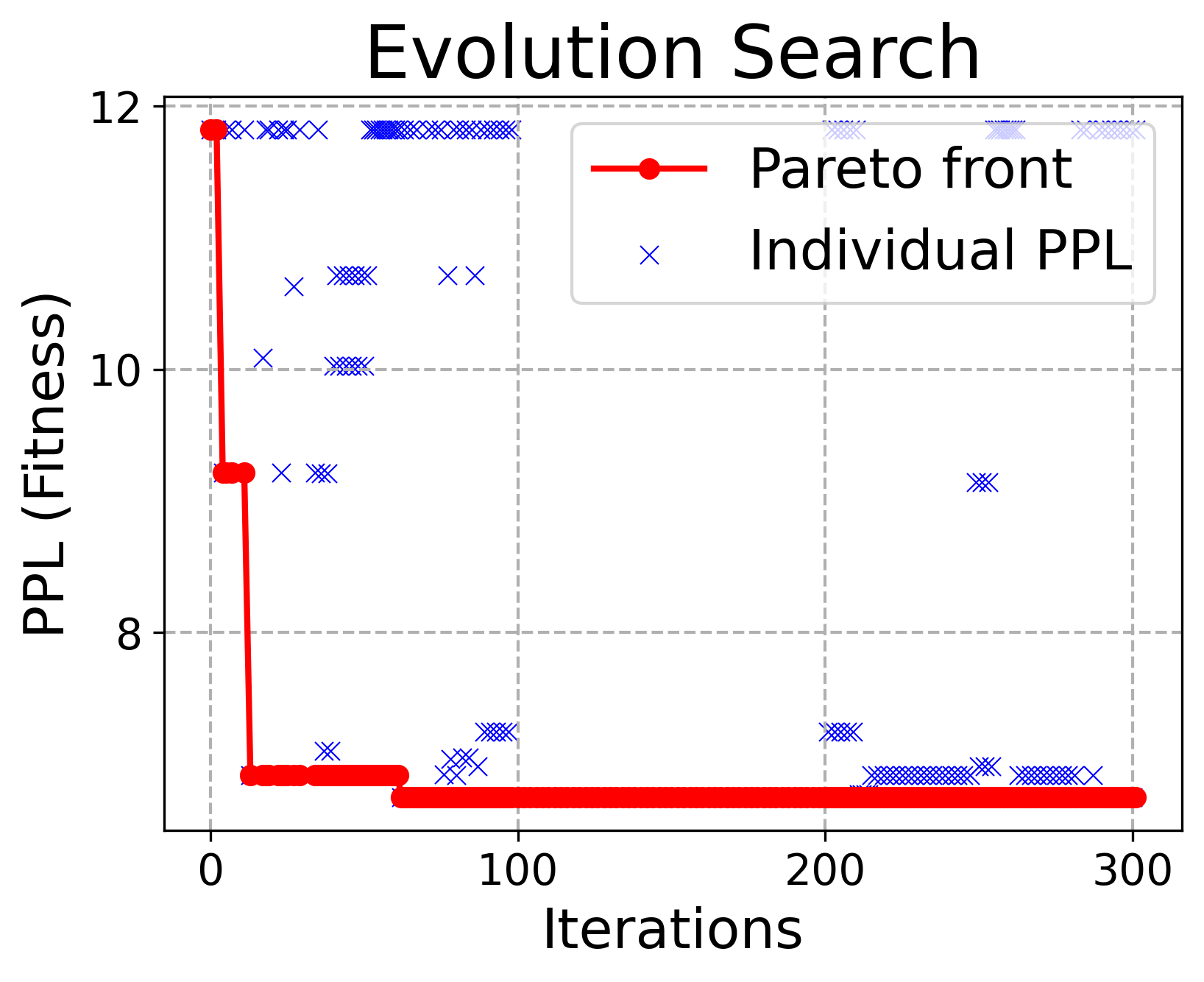}
        \label{fig:evolution-search}
    \end{minipage}
    \hfill
    \begin{minipage}[b]{0.49\linewidth}
        \centering
        \includegraphics[width=\linewidth]{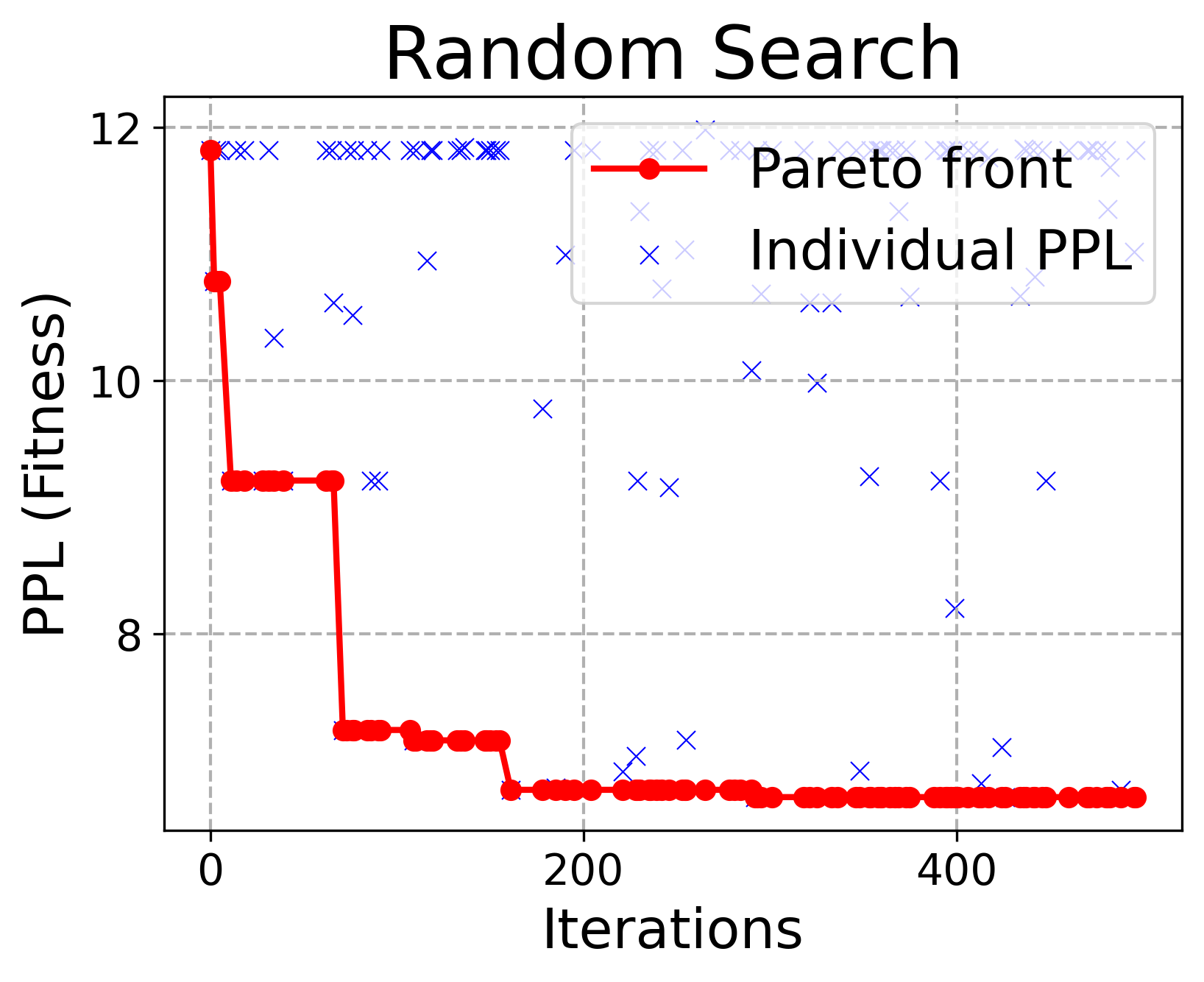}
        \label{fig:random-search}
    \end{minipage}
    \vskip -0.3in
    \caption{Comparison between Evolution Search and Random Search Processes. Notably, the individual perplexity is lower in the evolution search method, leading to significant improvements in the search efficiency and overall stability.}
    
    \label{ablation_evo_rnd_search}
\end{figure}
\begin{figure}[t]
    \centering
    \begin{minipage}{0.48\linewidth}
        \centering
        \includegraphics[width=\linewidth]{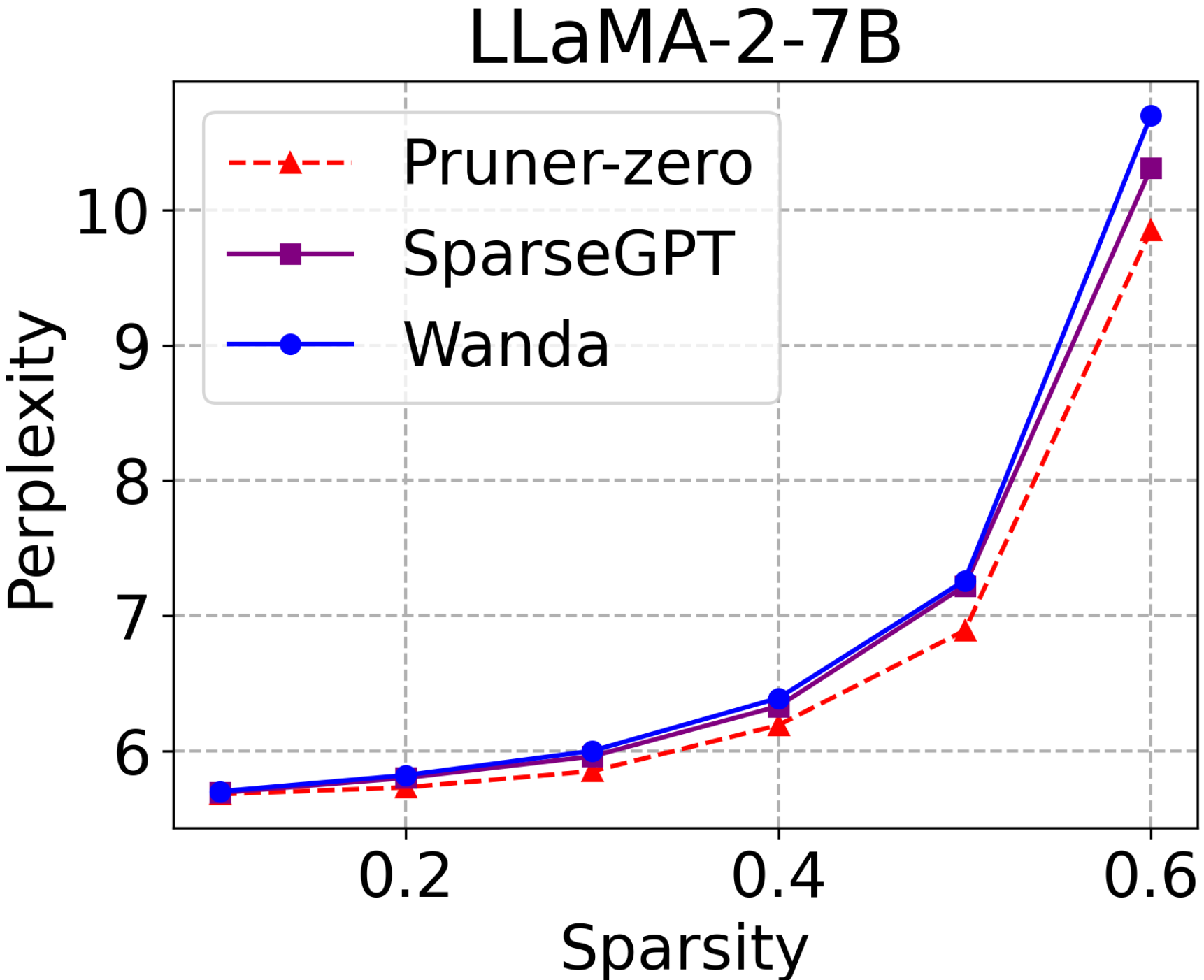}
        \label{fig:line-plot}
    \end{minipage}
    \hfill 
    \begin{minipage}{0.50\linewidth}
        \centering
        \includegraphics[width=\linewidth]{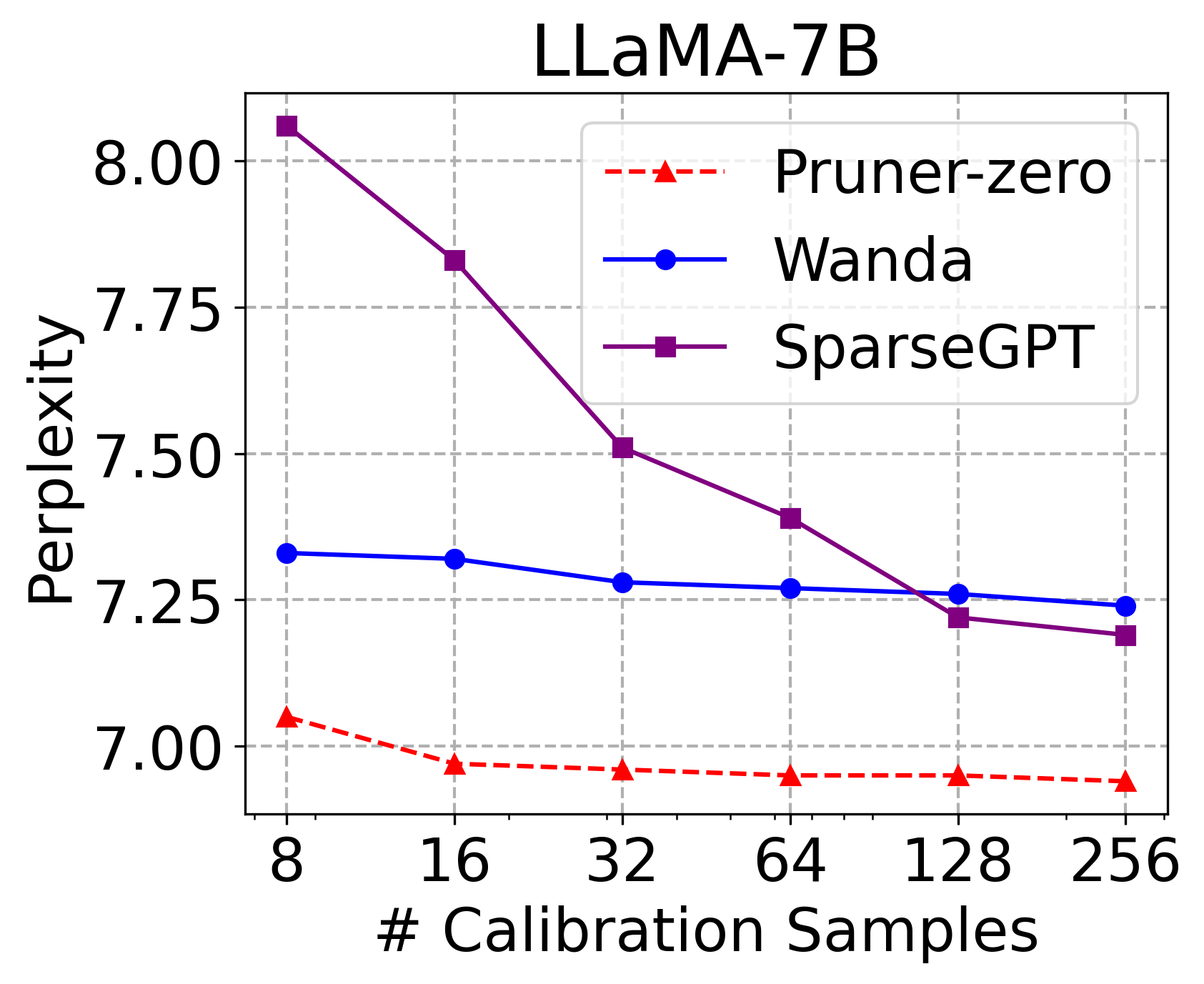}
        \label{fig:calibration-samples}
    \end{minipage}
    \vspace{-0.2in}
    \caption{Left: Perplexity under Various Sparsity Ratio; Right: Perplexity with Different Calibration Samples.}
    \label{fig:ablation_sparsity_calibration}
    \vskip -0.2in
\end{figure}

\noindent \textbf{Symbolic Metric Initialization.} Each symbolic tree $\mathcal{S}_i$ represents an analytical equation aimed at quantifying the saliency of weights. To ensure balanced complexity throughout the population, we initialize symbolic metrics with depths varying from three to five. Given the necessity for shape conformity between the symbolic tree and the weights, we impose specific restrictions on the activation because the shape of activation $X$ differs from that of weight $W$. Typically, the norm of activation $||X||_2$ is utilized for computation to address this discrepancy. To resolve the shape-matching issue, we stipulate that if $||X||_2$ serves as the child node of a unary operation, it is excluded from the search space, and a new node is generated in its stead. This process is illustrated in the \textit{SymbolicMetricInitialization} function in line 2 of Algorithm~\ref{alg:evolution}.

\begin{table*}[t]
\centering
\small
\caption{Perplexity of pruned LLaMA and LLaMA-2 models on WikiText2. Our \method{} outperforms SparseGPT and Wanda, achieving lower Perplexity without weight updates.}
\vskip 0.05in
\label{tab:wikitext-lang-modeling}
\setlength{\tabcolsep}{6.5pt}
\renewcommand{\arraystretch}{1.2}
\begin{tabular}{l c c c c c c c c c }
\toprule 
        &  &  & \multicolumn{4}{c}{\textbf{LLaMA}} & \multicolumn{3}{c}{\textbf{LLaMA-2}}\\
    \cmidrule(lr){4-7}\cmidrule(l){8-10}
   \textbf{Method} &  \textbf{Weight Update} & \textbf{Sparsity} & \textbf{7B} & \textbf{13B} & \textbf{30B} & \textbf{65B} & \textbf{7B} & \textbf{13B} & \textbf{70B} \\
    \midrule
      Dense  & - & 0$\%$ & 5.68 & 5.09 & 4.77 & 3.56 & 5.12 & 4.57 & 3.12\\
    \hline
    Magnitude~\cite{song2016deep_compression} & \xmark & 50$\%$ &17.29 & 20.21 & 7.54 & 5.90 & 14.89 & 6.37 & 4.98\\
    SparseGPT~\cite{Frantar2023SparseGPTML} & \cmark & 50$\%$ & 7.22 & 6.21 & 5.31 & 4.57 & 6.51 & 5.63 & 3.98\\
    Wanda~\cite{Sun2023ASA_wanda}     & \xmark & 50$\%$ & 7.26 & 6.15 & 5.24 & 4.57 & 6.42 & 5.56 & 3.98\\
  \gr \method{}& \xmark & 50$\%$ & \tbf{6.95} & \tbf{5.94} & \tbf{5.01} & \textbf{4.33} & \tbf{6.26} & \tbf{5.36} & \tbf{3.82} \\
    \hline
      Magnitude~\cite{song2016deep_compression} & \xmark & 4:8 & 16.84 & 13.84 & 7.62 & 6.36 & 16.48 & 6.76 & 5.54\\
      SparseGPT~\cite{Frantar2023SparseGPTML} & \cmark & 4:8 &  8.61 & 7.40 & 6.17 & 5.38 & 8.12 & 6.60 & 4.59\\
      Wanda~\cite{Sun2023ASA_wanda} & \xmark & 4:8 & 8.57 & 7.40 & 5.97 & 5.30 & 7.97 & 6.55 & 4.47\\
      \gr  \method{}& \xmark & 4:8 & \tbf{8.12} & \tbf{6.81} & \tbf{5.65} & \textbf{4.92} & \tbf{7.67} & \tbf{6.10} & \tbf{4.31}\\
    \hline
        Magnitude~\cite{song2016deep_compression} & \xmark & 2:4 & 42.13 & 18.37 & 9.10 & 7.11 & 54.59 & 8.33 & 6.33\\
        SparseGPT~\cite{Frantar2023SparseGPTML} & \cmark & 2:4 & 11.00 & 9.11 & 7.16 & 6.28 & \tbf{10.17} & 8.32 & 5.40\\
        Wanda~\cite{Sun2023ASA_wanda} & \xmark & 2:4 & 11.53 & 9.58 & 6.90 &  6.25 & 11.02 & 8.27 & 5.16\\
      \gr \method{}& \xmark & 2:4 & \tbf{10.61} & \tbf{8.11} & \tbf{6.51} &  \tbf{5.67} & 10.52 & \tbf{7.41} & \tbf{4.81}\\
    \bottomrule
\end{tabular}
\vskip -0.05in
\end{table*}

\begin{table*}[t]
\centering
\caption{Mean zero-shot accuracies ($\%$) of pruned LLaMA and LLaMA-2 models on the BoolQ, RTE, HellaSwag, WinoGrande, ARC, and OBQA datasets. Wanda performs competitively against prior best method SparseGPT, without introducing any weight update.  
}
\label{tab:zero_shot_results}
\renewcommand{\arraystretch}{1.2}
\small
\begin{tabular}{l c c c c c c c c c }
\toprule 
        & & & \multicolumn{4}{c}{\textbf{LLaMA}} & \multicolumn{3}{c}{\textbf{LLaMA-2}}\\
    \cmidrule(lr){4-7}\cmidrule(l){8-10}
   \textbf{Method} & \textbf{Weight Update} & \textbf{Sparsity} & \textbf{7B} & \textbf{13B} & \textbf{30B} & \textbf{65B} & \textbf{7B} & \textbf{13B} & \textbf{70B} \\
    \midrule
      Dense  & - & 0$\%$ & 59.99 & 62.59 & 65.38 & 66.97 & 59.71 & 63.03 & 67.08  \\
    \hline
    Magnitude~\cite{song2016deep_compression} & \xmark & 50$\%$ & 46.94 & 47.61 & 53.83 & 62.74 & 51.14 & 52.85 & 60.93 \\
    SparseGPT~\cite{Frantar2023SparseGPTML} & \cmark & 50$\%$ & 54.94 & 58.61 & 63.09 & 66.30 & 56.24 & 60.72 & 67.28  \\
    Wanda~\cite{Sun2023ASA_wanda} & \xmark     & 50$\%$ & 54.21 & 59.33 & 63.60 & 66.67 & 56.24 & 60.83 & 67.03 \\
  \gr \method{}& \xmark & 50$\%$ & \textbf{59.56} & \textbf{62.67} & \textbf{67.49} & \textbf{69.81} & \textbf{58.87} & \textbf{64.83} & \textbf{71.10} \\
    \hline
      Magnitude~\cite{song2016deep_compression} & \xmark & 4:8 & 46.03 & 50.53 & 53.53 & 62.17 & 50.64 & 52.81 & 60.28\\
      SparseGPT~\cite{Frantar2023SparseGPTML} & \cmark & 4:8 &  52.80 & 55.99 & 60.79 & 64.87 & 53.80 & 59.15 & 65.84\\
      Wanda~\cite{Sun2023ASA_wanda} & \xmark & 4:8 & 52.76 & 56.09 & 61.00 & 64.97 & 52.49 &  58.75 & 66.06 \\
      \gr  \method{}& \xmark & 4:8 & \textbf{56.24} & \textbf{59.03} & \textbf{64.04} & \textbf{68.04} & \textbf{55.82} &  \textbf{61.97} & \textbf{69.94} \\
    \hline
    Magnitude~\cite{song2016deep_compression} & \xmark & 2:4 & 44.73 & 48.00 & 53.16 & 61.28 & 45.58 & 49.89 & 59.95 \\
    SparseGPT~\cite{Frantar2023SparseGPTML} & \cmark & 2:4 & 50.60 & 53.22 & 58.91 & 62.57 & 50.94 & 54.86 & 63.89  \\
    Wanda~\cite{Sun2023ASA_wanda} & \xmark & 2:4 & 48.53 & 52.30 & 59.21 & 62.84 & 48.75 & 55.03 & 64.14 \\
    \gr \method{}& \xmark & 2:4 & \textbf{52.06} & \textbf{56.78} & \textbf{62.00} & \textbf{65.42} & \textbf{52.02} & \textbf{58.38} & \textbf{67.69} \\
    \bottomrule
\end{tabular}
\vspace{-1ex}
\end{table*}

\noindent \textbf{Crossover and Mutation.} As shown in Figure~\ref{fig:main_figure}, the process of subtree crossover involves the selection of two subtrees from the parent symbolic trees at random, followed by their subsequent interchange, as shown in the lines 8 and 9 of Algorithm~\ref{alg:evolution}. This crossover technique facilitates the combination of genetic symbols from parents, potentially leading to offspring with enhanced performance metrics. Following the crossover, node mutation is conducted with a probability of $p$. This mutation process is constrained to altering nodes within the same operation type, ensuring that unary operations can only mutate into other unary operations, and similarly for binary operations. This constraint maintains the structural integrity of the symbolic trees.

\noindent \textbf{Opposing Operation Simplification (OOS).} During our exploration of the search space, we observe that equivalent symbolic pruning metrics frequently encompass pairs of antagonistic operations, such as \textit{exp} and \textit{log}, or \textit{sub} and \textit{neg}, leading to an unnecessary increase in complexity. This redundancy detracts from the efficiency of the search process and obfuscates the interpretability of the expression trees. The OOS methodology is designed to refine the search space by eliminating these pairs of opposing operations, thereby streamlining the genetic representation and facilitating the identification of more concise and potent symbolic expressions. To implement this strategy, we meticulously catalog all opposing operations as detailed in Appendix~\ref{appendix:detailed_correlation_analysis}. Upon detecting operations within the same symbolic pruning metric that are opposed, we proceed to eliminate them as in line 17 of Algorithm~\ref{alg:evolution}.

\noindent \textbf{Perplexity as Fitness.} Upon generating offspring metrics, we undertake a post-training pruning evaluation using the LLaMA-2-7B model on the WikiText2 dataset~\cite{merity2017pointer_wikitext2} to determine their perplexity, which serves as the measure of fitness. In this context, a lower perplexity score indicates superior fitness, reflecting a more effective pruning metric in terms of model performance.

\noindent \textbf{Searched Symbolic Pruning Metric.} Here is the formula of the searched Symbolic Pruning Metric: 
\begin{equation}
    \text{Pruner-Zero}= \left|\left|W\right|\times \left|W\right|\right| \times \sigma(|G|)
\end{equation}\label{eq:searched_spm}
where $(\sigma)$ represents the min-max scaling function. This scaling function is pivotal in normalizing the weights $(\lvert W \rvert)$ and gradients $(\lvert G \rvert)$. By applying the min-max scaling, the weights and gradients are transformed to a common scale without distorting differences in the ranges of values. The squaring of the normalized weights and the subsequent multiplication with the normalized gradients, all under a square root, suggests a geometric mean approach. This form of averaging is beneficial for pruning metrics as it tends to dampen the effect of extreme values.

\section{Experiments}\label{experiments}

\subsection{Models and Implementation Details}

\textbf{Implementation Details.} 

In this section, we primarily assess the effectiveness of our \method{} on two of the most extensively used LLM families: LLaMA 7B/13B/30B/65B~\cite{touvron2023llama} and LLaMA-2-7B/13B/70B~\cite{touvron2023llama2}. To further explore the generalizability of \method, we apply the developed symbolic pruning metric to earlier LLM families, such as OPT~\cite{Zhang2022OPTOP} and Tiny-LLaMA~\cite{Zhang2024TinyLlamaAO}, as detailed in Appendix~\ref{sec:other_llm_families}. The performance of the pruned models is evaluated in terms of language modeling and zero-shot tasks. For language modeling, we follow the established protocols in LLM compression research~\cite{Sun2023ASA_wanda, Frantar2023SparseGPTML} to assess the perplexity on the WikiText2~\cite{merity2017pointer_wikitext2} validation set. For zero-shot tasks, we utilize seven tasks from the EleutherAI LM Harness~\cite{eval-harness}.

\noindent \textbf{Counterparts.} In comparing \method{} with established Post-training Pruning methods, magnitude pruning~\cite{Han2015LearningBW} serves as a baseline, removing weights based on the absolute value of magnitude. SparseGPT~\cite{Frantar2023SparseGPTML} introduces second-order pruning for LLMs, incorporating layer-wise reconstruction, while Wanda~\cite{Sun2023ASA_wanda} simplifies this by using a diagonal approximation, avoiding matrix inverse calculations. Our approach avoids the computationally expensive Hessian matrix and aligns with magnitude pruning and Wanda in forgoing retraining or weight updates during pruning. Unlike SparseGPT and Wanda, which require calibration data, we use WikiText2 to estimate input statistics like Gradients (G) and Activations (X).

\noindent \textbf{Sparsity.} We adopt a uniform sparsity across all linear layers in unstructured pruning settings, maintaining consistency with the approach used by Wanda. Our focus for all pruning methods is exclusively on the linear layers of the model. To ensure a fair comparison with SparseGPT and Wanda, we assess three types of sparsity: unstructured, structured 4:8, and structured 2:4. These sparsity types are chosen to reflect different levels of granularity and constraints in pruning. All our implementations are based on Wanda's framework, providing a robust and consistent basis for evaluating the effectiveness of the various pruning strategies.

\noindent \textbf{Evolution Settings.} The evolutionary search starts with an initial population of 50 and 300 iterations. The depth of symbolic trees ranges from 3 to 5. Tournament selection utilizes a top-$K$ parameter of 10, selecting two parent symbolic pruning metrics from the 10 best-performing candidates. The mutation probability is set to 0.5. This search, focused on identifying an optimal symbolic pruning metric, is executed using the LLaMA-2-7B model. Perplexity is evaluated under unstructured pruning with 50\% sparsity. The searched metric, detailed in Table~\ref{tab:property}, is applied to various model sizes. The genetic programming tasks are executed on two NVIDIA 4090 GPUs, while the generalization experiments involving zero-shot tasks and language modeling on the LLaMA-2-70B are conducted using 8 A100 GPUs.

\begin{table*}[t]
\centering
\caption{Perplexity of the pruned OPT family models with various sizes on WikiText2. $\dag$ denotes updating weights using 128 samples. $\S$ denotes we utilize 128 samples using iterative update.}\label{tab:opt_family}
\renewcommand{\arraystretch}{1.2}
\resizebox{.75\textwidth}{!}{
\begin{tabular}{lccccccccc}
\toprule 
        & & & \multicolumn{6}{c}{\textbf{OPT}} \\
    \cmidrule(lr){4-9}
   \textbf{Method} & \textbf{Weight Update} &\textbf{Sparsity} & \textbf{125m} & \textbf{350m}  & \textbf{1.3B} & \textbf{2.7B} & \textbf{6.7B} & \textbf{13B} \\ 
    \midrule
    Dense  & - & 0$\%$ & 27.66 & 22.00 & 14.62 & 12.47 & 10.86 & 10.13\\
    \hline
    Magnitude & \xmark & 50$\%$ & 7e3 & 6e3 & 1e4 & 9e3 & 9e4 & 2e4 \\
    Wanda & \xmark & 50$\%$  & 38.96 & 35.92 & 19.12 & 14.28 & 11.94 & 11.42 \\
    \gr \method{}& \xmark & 50$\%$  & \textbf{37.69} & \textbf{35.91} & \textbf{18.19} & \textbf{13.85} & \textbf{11.86} & \textbf{11.32} \\ \hline 
    SparseGPT & \cmark & 50$\%$ &  37.07 & 34.76 & \textbf{17.44} & 13.48 & 11.57 & 11.19 \\
    \gr \method{}$^\dag$ & \cmark & 50$\%$ & \textbf{34.04}  & 30.47  & 18.57  & 13.31  & \textbf{11.52}  & 11.54  \\
    \gr \method{}$^\S$ & \cmark & 50$\%$ & 35.51  & \textbf{29.75}  & 18.60  & \textbf{13.24}  & \textbf{11.52}  & \textbf{10.86}  \\
    \bottomrule
\end{tabular}}
\vskip -0.15in
\end{table*}

\begin{table}[t]
    \centering
    \caption{Perplexity and Searched SPM w.r.t. OOS}
    \label{tab:ablation_oss}
    \begin{tabular}{ccc}
    \toprule 
    \textbf{Method}    & \textbf{Perplexity} & \textbf{Searched SPM Expression} \\ \midrule
    w OOS     & 6.7079     & $||W||\times ||W|| \times \sigma(G)$ \\
    w/o OOS   & 6.8395     & $\frac{||W||+G}{\sqrt{||W||}}+\sigma(\zeta(||W||))$ \\ 
    \bottomrule
    \end{tabular}
\end{table}

\begin{table}[t]
\renewcommand{\arraystretch}{0.9}
\caption{Fine-tuning can mitigate the perplexity gap to dense LLM.}
\label{tab:finetune_results}
\resizebox{\columnwidth}{!}{
    \begin{tabular}{lccccc}
    \toprule 
        \textbf{Evaluation} &   \textbf{Dense} & \textbf{Fine-tuning} &  \textbf{50\%} & \textbf{4:8} & \textbf{2:4} \\
        \midrule
        \gr \wc    \multirow{2}{*}{LLaMa-7B} & \wc \multirow{2}{*}{5.68} & \xmark & 6.95 & 8.12 & 10.61 \\
                    & & LoRA & 6.74 & 7.40 & 8.24 \\
         \hline
       \gr \wc    \multirow{2}{*}{LLaMa-2-7B} & \wc \multirow{2}{*}{5.12} & \xmark & 6.26 & 7.67 & 10.52  \\
              & & LoRA & 6.41 & 7.01 & 7.73 \\
        \bottomrule
    \end{tabular}
}
\vskip -0.1in
\end{table}

\subsection{Language Modeling}

As shown in Table~\ref{tab:wikitext-lang-modeling}, this study reports on the perplexity of pruned LLaMA and LLaMA-2 models under 50\% unstructured and 2:4, 4:8 structured pruning scenarios. We annotate the necessity of weight updates and the corresponding sparsity levels for both unstructured and structured pruning methods.
Remarkably, \method{} outperforms all established pruning techniques without requiring any weight updates. Notably, it surpasses the magnitude baseline by a significant margin. A trend observed is that \method{} demonstrates a lesser performance drop in larger ones. For instance, in unstructured pruning, the LLaMA-7B model shows a decrease of 1.27 in perplexity, whereas the LLaMA-30B model only exhibits a 0.24 drop. Similarly, the LLaMA-2-7B model experiences a 1.14 drop in perplexity, while the drop for the LLaMA-2-70B model is only 0.7. This indicates that \method{} is particularly advantageous for larger models, such as LLaMA-30B and LLaMA-2-70B. In comparison, \method{} consistently outperforms Wanda and mostly surpasses SparseGPT, except for the LLaMA-2 model in the 2:4 structured pruning setting.
For robustness analysis under different seeds, refer to Table~\ref{appendix:tab:seed}. For experiments on previous pruning methods applied to BERT~\cite{bert}, see Appendix~\ref{appendix:previous_pruning_methods}. For comparison with GBLM-Pruner~\cite{Das2023BeyondSH_GBLM}, see Table~\ref{appendix:tab:gblm-pruner}.

\subsection{Zero-shot Tasks} 

To assess the generalizability of \method{}, we evaluate its performance on seven common-sense tasks from the Eleuther AI lm-evaluation-harness benchmark~\cite{eval-harness} in a zero-shot setting. The selected tasks encompass BoolQ~\cite{Clark2019BoolQET}, RTE~\cite{Wang2018GLUEAM}, HellaSwag~\cite{Zellers2019HellaSwagCA}, WinoGrande~\cite{Sakaguchi2019WinoGrande}, ARC~\cite{Clark2018ThinkYH}, and OBQA~\cite{Mihaylov2018CanAS}. Although zero-shot evaluation on individual tasks can exhibit variability, Table~\ref{tab:zero_shot_results} presents the average performance of \method{} across all seven tasks to enhance the interpretability of the model's overall performance.

The results demonstrate that our \method{} significantly outperforms the magnitude baseline and competes effectively with previous top-performing approaches, such as SparseGPT and Wanda. Additionally, we observe that with increasing model size, the accuracy gap in zero-shot tasks decreases. The larger models like LLaMA-2-70B even surpass the dense model baselines. Detailed performance metrics for each task are provided in Appendix~\ref{appendix:zero_shot}.

\subsection{Evaluation of In-Context Learning}

In-context learning, a pivotal capability of large language models (LLMs), is crucial for tasks requiring adaptability and reasoning without explicit retraining. To assess this capability within various models, we focused our evaluation on the GSM8K dataset, comprised of diverse and complex grade school math problems that effectively challenge the models' reasoning abilities. Utilizing the LightLLM framework~\cite{lightllm}, a Python-based tool optimized for efficient and scalable inference with LLMs, we tested the LLaMA2 13B model's in-context learning performance using a fixed set of eight demonstrations aligned with the Chain-of-Thought~\cite{wei2022chain} approach. The comparative analysis presented in Table~\ref{tab:in_context_learning} indicates that Pruner-Zero outperforms other pruning approaches such as SparseGPT~\cite{Frantar2023SparseGPTML} and Wanda~\cite{Sun2023ASA_wanda} in the in-context learning tasks, despite the general trend of decreased performance post-pruning. This observation, consistent with findings from recent studies~\cite{Li2024EvaluatingQL} documenting a reduction in in-context learning capabilities of LLMs following quantization, highlights a significant challenge in the field of model pruning and optimization: preserving the nuanced capabilities of LLMs while reducing their computational overhead. Although Pruner-Zero shows a marked improvement over other methods, the noticeable decline in the performance of the unpruned (Dense) model underscores the delicate balance between model size and functionality, particularly in tasks requiring high cognitive functions such as reasoning and comprehension.

\begin{table}[t]
\centering
\small 
\caption{In-Context Learning Accuracy on the GSM8K Dataset}
\label{tab:in_context_learning}
\vskip 0.15in
    \begin{tabular}{l|c}
    \toprule
    \textbf{Dataset} & GSM8K \\
    \midrule
    Dense & 0.287 \\
    \midrule 
    Magnitude & 0.0607 \\
    SparseGPT~\cite{Frantar2023SparseGPTML} & 0.1152 \\
    Wanda~\cite{Sun2023ASA_wanda} & 0.1312 \\
    Pruner-Zero & 0.1403 \\
    \bottomrule
    \end{tabular}
\end{table}

\subsection{Ablation Study}

\textbf{Evolution Search vs. Random Search.} To evaluate the efficacy of Evolution Search, ablation studies were conducted, the results of which are presented in Figure~\ref{ablation_evo_rnd_search}. It was observed that Evolution Search achieves convergence in fewer than 100 iterations, in contrast to Random Search, which requires approximately 300 iterations to converge.

\textbf{Robustness across Different Sparsity Ratios.} The left side of Figure~\ref{fig:ablation_sparsity_calibration} presents the perplexity results for LLaMA-2-7B on WikiText2 across varying sparsity ratios, ranging from 0.1 to 0.6. These results demonstrate that \method{} consistently outperforms SparseGPT and Wanda across all tested sparsity levels. This confirms the robustness and effectiveness of our symbolic pruning metric under diverse sparsity ratios, ensuring reliable performance even as the level of pruning increases.

\textbf{Robustness to Calibration Sample Size.} The right side of Figure~\ref{ablation_evo_rnd_search} illustrates the impact of varying calibration sample sizes, ranging from 8 to 256 samples. A distinct trend emerges as the calibration sample size changes: all pruning methods demonstrate increased robustness and lower perplexity. Our \method{} demonstrates robustness to calibration samples; specifically, when the sample size exceeds 16, the perplexity value is below 7. Finally, we chose 128 as the calibration size to ensure alignment with Wanda.

\textbf{Robustness to OPT family.} Table~\ref{tab:opt_family} showcases the pruning performance on OPT models ranging from 125 million to 13 billion parameters. We find that Pruner-Zero outperforms both Magnitude and Wanda in post-training pruning without weight updates across various model sizes. Our Pruner-Zero also outperforms SparseGPT with weight updates following Wanda~\cite{Sun2023ASA_wanda}. A sequential update ($\dag$) means that at each layer, the full pruned mask is first computed and a weight update is performed on the remaining weights. An iterative update ($\S$) means that the pruning and weight update steps proceed iteratively within each layer.

\textbf{Effectiveness of OSS.} Opposing Operation Simplification (OOS) strategy plays a crucial role in reducing redundancy among the candidate symbolic pruning metrics. Without OOS, the population may contain a large number of mathematically equivalent metrics, which is undesirable. The OOS strategy helps maintain diversity in the candidate solution space by eliminating such redundancies, aligning the search process with our expectations, and improving its efficiency.
To quantize the influence of OOS, we conducted experiments under different depths of the symbolic tree using Effective probability. Effective probability calculates the frequency of when OOS works. The effective probability is computed based on 1,000 randomly generalized trees. And we conduct four trials to get the final result. From Table ~\ref{tab:redundency_in_ss}, we can find that the effective probability is very high denoting the redundancy of the search space. Besides, we present the searched SPM and their corresponding perplexity w.r.t OOS strategy in Table~\ref{tab:ablation_oss}. The experiments were conducted for 300 iterations on the LLaMA-2-7B model with 50\% unstructured pruning settings. The experiments without the OOS strategy showed deteriorated performance, likely due to redundancy in the small population size of 50, where high homogeneity in metrics prevented further performance increase.

\subsection{Analysis}\label{sec:analysis}

\textbf{Correlation Analysis.} During the evolutionary search, we gathered various symbolic pruning metrics and their corresponding perplexities, identifying those with perplexities below 7 as potential metrics (detailed in Table~\ref{appendix:tab:equation_searched}, Appendix~\ref{appendix:searched_pruning_metric}). The analysis of operation frequencies in these metrics, as shown in Figure~\ref{fig:correlation_between_operations}, revealed several key insights: (1) Multiplication operations exhibit a strong correlation with perplexity, highlighting their importance. (2) Opposite operations (\textit{sqrt} and \textit{sqr}) show a high correlation, suggesting counteractive effects. (3) The min-max scaling (\textit{mms}) operation is robust, displaying a low correlation with other operations, underscoring its significance. These observations support our metric development discussed in Section~\ref{genetic_programming_framework}.

\textbf{LoRA Fine-tuning.} We further explore the potential of fine-tuning to mitigate the performance reduction observed in pruned Large Language Models (LLMs). Specifically, we employ LoRA~\cite{hu2022lora} ($r=8$) for fine-tuning on the C4 training dataset~\cite{2019t5_c4dataset} using 1 GPU and 12 hours, targeting the auto-regressive loss. Experiments were conducted on both the LLaMA and LLaMA-2-7B models, encompassing 50\% unstructured as well as 2:4 and 4:8 structured pruning scenarios. The resulting perplexity on WikiText2 is detailed in Table~\ref{tab:finetune_results}. In most cases, fine-tuning with LoRA can restore the performance of pruned LLMs, especially for hard cases like 2:4 structured pruning.

\section{Conclusion}

In this paper, we present a \method{} framework for symbolic pruning metric discovery for Large Language Models (LLMs) by formulating it as a symbolic regression problem. We leverage genetic programming to efficiently search for superior symbolic pruning metrics that have lower perplexity after pruning. During the search, we find the opposing operation that affects the quality of the population and propose the Opposing Operation Simplification (OOS) strategy to enhance search efficiency. Our comprehensive experimental evaluation, conducted on the LLaMA and LLaMA-2 for both language modeling and zero-shot tasks, reveals that our \method{} framework surpasses current state-of-the-art (SOTA) methods, including Wanda and SparseGPT, in terms of both structured and unstructured pruning

\section*{Acknowledgements}

This work was partially supported by a Hong Kong RIF grant under Grant No. R6021-20, and Hong Kong CRF grants under Grant No. C2004-21G and C7004-22G.

\section*{Impact Statement}

\method{} significantly refines the post-training pruning process of large language models, employing genetic programming to evolve symbolic pruning metrics. This method effectively identifies and eliminates non-critical parameters, thereby preserving the model's performance. The focus of this research is on technological innovation, and it does not extend to the analysis of ethical considerations or societal impacts.

\normalem
\bibliography{citations}
\bibliographystyle{icml2024}

\newpage
\appendix
\onecolumn

\section{Related Work}

The development of Large Language Models (LLMs) has witnessed a surge in model and dataset sizes, necessitating distributed training across numerous devices \citep{tang2020survey,tang2023fusionai,tang2024fedimpro}. This distributed approach, while effective, demands substantial computational and storage resources, with LLMs incurring higher energy costs compared to their smaller counterparts \citep{luccioni2023estimating,schwartz2020green,tangDVFS}. Consequently, energy-efficient LLM training and inference are crucial for green computing, with LLM pruning emerging as a key technique for achieving this goal. Post-training pruning, in particular, has gained prominence due to its minimal resource requirements, making it a cost-effective approach for democratizing access to LLMs \citep{miao2022learning, Frantar2023SparseGPTML, Sun2023ASA_wanda}. This method's efficiency and accessibility contribute significantly to the broader impact and applicability of LLMs.

\paragraph{Network Pruning} Network pruning is an effective technique for reducing model complexity while preserving performance, although it often requires extensive retraining. However, traditional pruning methods~\cite{hoang2023revisiting, sreenivasan2022rare, liu2018rethinking, chen2023otov, NEURIPS2021_23e582ad} become impractical when dealing with the substantial parameter sizes and vast datasets of Large Language Models (LLMs). Deep Compression~\cite{han2015deep_compression_magnitude} popularized magnitude-based pruning for deep neural networks, which removes the weights with the smallest absolute values, assuming that they have the least impact on the network's output. Network pruning techniques can be broadly categorized into two main approaches: unstructured pruning and structured pruning.

\begin{table}[b!]
\centering
\small 
\caption{Comparison of Pruner-Zero with its counterparts.}
\label{tab:ml_optimization}
\vskip 0.15in

\begin{tabularx}{\textwidth}{l|XXXXX}
\toprule 
\textbf{Description} & \textbf{AutoML-Zero~\cite{real2020automlzero}} & \textbf{EZNAS~\cite{akhauri2022eznas}} & \textbf{Auto-Prox\cite{wei2024auto}} & \textbf{EMQ~\cite{dong2023emq}} & \textbf{Pruner-Zero} \\ \midrule
\textbf{Task} & Machine Learning Program Discovery & Zero-shot NAS & Zero-shot NAS & Mixed-precision Quantization & Symbolic Pruning Metric \\ \hline
\textbf{Targets} & - & CNN & ViT & CNN & LLMs \\ \hline
\textbf{Params} & - & 0.3-1.5MB & 2-25MB & 13.4-44.6MB & 7B-70B \\ \hline
\textbf{Retrain} & No & Yes & Yes & Yes & No \\ \hline
\textbf{Input} & Multiple inputs & One or two inputs & Two inputs & Two inputs & Unlimited inputs \\ \hline
\textbf{Output} & Machine Learning Task (accuracy) & One scalar & One scalar & One scalar & Matrix with the same shape as weight \\ \hline
\textbf{Strategy} & Evolutionary Algorithm & Distributed Evolutionary Algorithm in Python (DEAP) & Elitism-Preserve Strategy & Diversity Prompting Selection & Opposing Operation Simplification strategy \\ \hline
\textbf{Objective} & Find machine learning algorithms from scratch & Find optimal proxy that can measure the convolution-based architectures & Find the optimal proxy that can measure the vit-based architectures & Find the optimal metric that can better rank candidate bit-width configurations & Find the optimal symbolic pruning metric that can measure the importance of different weights \\ 
\bottomrule
\end{tabularx}
\end{table}

\paragraph{(1) Unstructured Pruning} involves removing individual weights or connections based on certain criteria. 
SparseGPT~\citep{Frantar2023SparseGPTML} is the first post-training quantization method that performs unstructured pruning using an approximated Hessian matrix. Wanda~\citep{Sun2023ASA_wanda} further simplifies the Hessian matrix by using just the weight and $l_2$ norm of activation. GBLM-Pruner~\citep{Das2023BeyondSH_GBLM} further introduces the gradient to boost the performance. Plug-and-play~\cite{zhang2024plugandplay} The paper presents a plug-and-play post-training pruning method for large language models (LLMs) that introduces two innovative components: Relative Importance and Activations (RIA), a new pruning metric, and Channel Permutation, a technique to maximize the preservation of important weights under N:M sparsity constraints. The proposed method, named plug-and-play, outperforms existing pruning techniques and achieves practical speed-up on specific hardware without the need for additional fine-tuning or retraining.  PERP~\citep{Zimmer2023PERPRT} uses Low-rank adaptation to mitigate the expense of the retraining process in the original prune-retrain paradigm. NutePrune~\citep{Li2024NutePruneEP} combines structure pruning with progressive knowledge distillation by utilizing the unpruned model as a teacher and the pruned model as a student. OWL~\citep{yin2024outlier_owl} proposed outliers metric to re-assign the sparsity of different layers. BESA~\citep{xu2024besa} proposes to use parameter-efficient sparsity learning to learn the sparsity ratio in a differentiable manner. GRAIN~\cite{Yang2022GradientbasedIP} utilizes gradient information to prune intra-attention structures, incorporating knowledge distillation to enhance performance.

\paragraph{(2) Structured Pruning}, also known as N:M structured pruning, is a specialized technique designed to enhance computational efficiency during the inference phase in deep learning models. This method is particularly tailored for compatibility with NVIDIA’s Ampere architecture and its Sparse Tensor Cores. In the context of N:M structured pruning, for each block of M parameters, only N parameters are actively retained while the remainder is pruned, effectively set to zero or eliminated. Common configurations such as 2:4 and 4:8 structured pruning maintain a fixed sparsity ratio of 50\%, indicating that only half of the parameters are preserved. Notably, the 4:8 structured pruning format offers less restrictive conditions than the 2:4 format, permitting a more adaptable parameter reduction. This flexibility in 4:8 structured pruning is demonstrated by superior performance metrics, as detailed in Tables 2 and 3, where 4:8 pruning consistently outperforms 2:4 configurations under identical experimental conditions. Furthermore, NVIDIA’s A100 GPU represents a significant milestone as the first mainstream hardware to incorporate sparse capabilities directly into its architecture. This allows the Sparse Tensor Cores of the A100 to support a wide array of operations prevalent in modern neural networks, including linear transformations, convolutional layers, recurrent neural networks, and transformer architectures. This integration heralds a new era of efficiency, enabling more rapid and energy-efficient computation across diverse deep-learning applications.
SliceGPT~\cite{slicegpt_iclr24}, focuses on compressing LLMs by eliminating rows and columns from weight matrices, though it is mainly effective at lower sparsity ratios. LLM-Pruner~\cite{Ma2023LLMPrunerOT} is inspired by DepGraph~\cite{Fang2023DepGraphTA} to detect the dependency lies in the model, which also considers both first-order information and approximated Hessian information.

\begin{table}[t]
\centering
\caption{Impact of Pruning Metrics on GBLM-Pruner Performance. This figure illustrates the performance variations under 50\% sparsity ratio in GBLM-Pruner \cite{Das2023BeyondSH_GBLM} when utilizing weight, gradient, and activation-based pruning metrics. The observed sensitivity of the results to seemingly minor format changes highlights the challenges of manual metric selection and underscores the need for an automated pruning framework, which serves as the motivation for this work.}
\vskip 0.15in
\label{tab:gblm-pruner-table}
\begin{tabular}{lc|lc}
\toprule 
\multicolumn{1}{c}{\textbf{Metric}} & \textbf{Perplexity} & \multicolumn{1}{c}{\textbf{Metric}} & \textbf{Perplexity} \\
\midrule
$|\mathbf{W}| \cdot ||G||_1$ & 7.17 & $(|\mathbf{W}| \cdot ||X||_2)^2 + \lambda \cdot |\mathbf{W}| \cdot ||G||_1$ & 180490.19 \\
$|\mathbf{W}| \cdot ||G||_2$ & 7.09 & $(|\mathbf{W}| \cdot ||X||_2)^2 + \lambda \cdot |\mathbf{W}| \cdot ||G||_2$ & 91781.49 \\
$|\mathbf{W}| \cdot ||X||_2 \cdot ||G||_1$ & 7.31 & $(|\mathbf{W}| \cdot ||X||_2)^2 - \lambda \cdot |\mathbf{W}| \cdot ||G||_1$ & 248646.28 \\
$|\mathbf{W}| \cdot ||X||_2 \cdot ||G||_2$ & 7.31 & $(|\mathbf{W}| \cdot ||X||_2)^2 - \lambda \cdot |\mathbf{W}| \cdot ||G||_2$ & 283620.75 \\
$|\mathbf{W}| \cdot ||X||_2^2 + \lambda \cdot |\mathbf{W}| \cdot ||G||_1$ & 6.86 & $(|\mathbf{W}| \cdot ||X||_2)^2 + \lambda \cdot |\mathbf{W}| \cdot ||G||_1$ & 6.90 \\
$|\mathbf{W}| \cdot ||X||_2^2 + \lambda \cdot |\mathbf{W}| \cdot ||G||_2$ & 6.89 & $(|\mathbf{W}| \cdot ||X||_2)^2 + \lambda \cdot |\mathbf{W}| \cdot ||G||_2$ & 6.88 \\
$|\mathbf{W}| \cdot ||X||_2^2 - \lambda \cdot |\mathbf{W}| \cdot ||G||_1$ & 1180.67 & $(|\mathbf{W}| \cdot ||X||_2)^2 - \lambda \cdot |\mathbf{W}| \cdot ||G||_1$ & 6.94 \\
$|\mathbf{W}| \cdot ||X||_2^2 - \lambda \cdot |\mathbf{W}| \cdot ||G||_2$ & 7.10 & $(|\mathbf{W}| \cdot ||X||_2)^2 - \lambda \cdot |\mathbf{W}| \cdot ||G||_2$ & 9377.00 \\
\bottomrule
\end{tabular}
\end{table}

\begin{table}[t]
\centering
\caption{Operations vocabulary. Summary of Unary and Binary Operations in Computational Functions. This table includes operation identifiers, code examples, input and output types, and detailed descriptions for each operation.}
\vskip 0.15in
\label{appendix:operation_vocabulary}
\resizebox{.9\textwidth}{!}{
\begin{tabular}{clcccl}
\toprule
\textbf{Op ID} & \textbf{Code Example}  & \textbf{Symbols}     & \textbf{Input / Types} & \textbf{Output / Type} & \textbf{Description}               \\ \midrule
U01            & $s_c$=sqr($s_a$)       & $(\cdot)^2$            & a / matrix                             & c / matrix                           & $s_c = s_a^2$                       \\
U02            & $s_c$=neg($s_a$)       & $-(\cdot)$             & a / matrix                             & c / matrix                           & $s_c = -s_a$                        \\
U03            & $s_c$=abs($s_a$)       & $|\cdot|$             & a / matrix                             & c / matrix                           & $s_c = |s_a|$                       \\
U04            & $s_c$=log($s_a$)       & $\text{log}$        & a / matrix                             & c / matrix                           & $s_c = \log(s_a)$                   \\
U05            & $s_c$=exp($s_a$)       & $e^{(\cdot)}$         & a / matrix                             & c / matrix                           & $s_c = e^{s_a}$                     \\
U06            & $s_c$=sqrt($s_a$)     & $\sqrt{\cdot}$      & a / matrix                             & c / matrix                           & $s_c = \sqrt{s_a}$                  \\
U07            & $s_c$=tanh($s_a$)  &  $\text{tanh}$   & a / matrix                             & c / matrix                           & $s_c = \tanh(s_a)$                  \\
U08            & $s_c$=pow($s_a$, $s_b$) & $(\cdot)^{(\cdot)}$    & a,b / matrixs                          & c / matrix                           & $s_c = s_a^{s_b}$                   \\
U09            & $s_c$=skp($s_a$)  & $\emptyset$             & a / matrix                             & c / matrix                           & $s_c = s_a$ (skip)       \\
U10            & $s_c$=mms($s_a$)  & $\sigma(\cdot)$                & a / matrix                             & c / matrix                           & $s_c = \text{min-max scale}(s_a)$   \\
U11            & $s_c$=zsn($s_a$)  & $\zeta(\cdot)$                 & a / matrix                             & c / matrix                           & $s_c = \text{z-score scale}(s_a)$   \\
U12            & $s_c$=norm$_2$($s_a$) & $||\cdot||_2$             & a / matrix                         & c / matrix                                & $s_c = ||s_a||_2$  \\
U13            & $s_c$=norm$_1$($s_a$) & $||\cdot||_1$             & a / matrix                         & c / matrix                                & $s_c = ||s_a||_1$  \\ \hline
B01            & $s_c$=add($s_a$, $s_b$) & $(\cdot)+(\cdot)$             & a,b / matrixs                          & c / matrix                           & $s_c = s_a + s_b$                   \\
B02            & $s_c$=sub($s_a$, $s_b$) & $(\cdot)-(\cdot)$              & a,b / matrixs                          & c / matrix                           & $s_c = s_a - s_b$                   \\
B03            & $s_c$=mul($s_a$, $s_b$) & $(\cdot)\times(\cdot)$       & a,b / matrixs                          & c / matrix                           & $s_c = s_a \cdot s_b$               \\
B04            & $s_c$=div($s_a$, $s_b$) & $(\cdot)/(\cdot)$             & a,b / matrixs                          & c / matrix                           & $s_c = \frac{s_a}{s_b}$             \\ 
\bottomrule
\end{tabular}}
\end{table}

\paragraph{Neural Architecture Search}~\cite{nasnet,ENAS} aims to automate the design of neural network architectures. Traditional methods~\cite{hu2021improving,linas2,li2023auto,li2024kd} for designing neural networks rely heavily on human expertise and extensive trial-and-error, which can be both time-consuming and resource-intensive. NAS methods~\cite{dong2023emq,lu2024uniads,dong2024parzc,He2022NASLIDEN} seek to alleviate this burden by leveraging algorithmic strategies to discover optimal network architectures. Our method is partly related to Zero-Shot Neural Architecture Search (Zero-Shot NAS)~\cite{wei2023tvt,Dong2023diswot,zhu2024saswot}, an emerging subfield within NAS. Zero-Shot NAS aims to predict the performance of neural network architectures without the need for extensive training and evaluation. The key to Zero-Shot NAS is the use of proxy metrics. Some Zero-Shot NAS methods propose to formulate the proxy as an automated discovery process~\cite{akhauri2022eznas, wei2024auto}. For example, EZNAS~\cite{akhauri2022eznas} and Auto-Prox~\cite{wei2024auto} represent significant advancements in the field of zero-shot NAS by focusing on the automatic identification of zero-cost proxies tailored to CNN-based and ViT-based architectures, respectively. These methodologies align with the broader domain of program synthesis, which is dedicated to the automated generation of programs that meet specific user-defined constraints. An illustrative example within this domain is AutoML-Zero~\cite{real2020automlzero}, which innovatively evolves complete machine learning algorithms from a foundational level.
Similarly, EMQ~\cite{dong2023emq} addresses the challenges of mixed-precision quantization. It aims to create a proxy capable of effectively ranking candidate bit-width configurations, thereby enhancing computational efficiency and model performance without extensive manual tuning.

\textbf{Characteristics and Innovations:}

\begin{itemize}
  \item \textbf{Model Scale:} Pruner-Zero is designed to manage exceptionally large-scale language models (LLMs), with parameters ranging from 7 billion to 70 billion. This scale surpasses that of the other discussed methods, highlighting its capability to handle extensive computational loads and complex data structures.
  \item \textbf{Retraining Necessity:} Unlike other approaches that necessitate retraining to fulfill their objectives, Pruner-Zero operates efficiently without the requirement for further retraining. This feature significantly reduces the computational overhead and accelerates the optimization process.
  \item \textbf{Input/Output Paradigms:} Pruner-Zero employs a unique approach to inputs and outputs; it accepts an unlimited number of inputs and produces an output matrix that retains the same dimensions as the model weights. This distinct I/O schema facilitates direct interventions in model pruning and optimization.
  \item \textbf{Objective:} The primary goal of Pruner-Zero is to discover an optimal symbolic pruning metric that assesses the significance of different weights in large language models. This objective focuses on enhancing model efficiency and performance through targeted weight reduction.
\end{itemize}

These distinctions underscore the varied landscape of program synthesis, where different problems are defined and differentiated by their specific constraints, which are inherently linked to the tasks at hand. Such innovations contribute substantially to the field, pushing the boundaries of what is possible in automated machine learning and architecture search.

\section{Motivation of Searching Pruning Metric}

In this section, our exploration into the design of an effective pruning metric is inspired by the GBLM-Pruner~\cite{Das2023BeyondSH_GBLM}. The intricacies of pruning metrics are exemplified in Table~\ref{tab:gblm-pruner-table}, which is derived from their paper. A meticulous analysis of these metrics underscores the complexity and iterative nature of their design process. Notably, metrics that are nearly identical in their formulation can yield vastly divergent outcomes, as evidenced by the last two metrics in the second column of the table. These metrics differ only slightly in their normalization approach, yet the resulting perplexities are starkly contrasting (6.94 vs. 9377). Such findings illuminate the challenges faced in metric design and catalyze for our development of \method, an automated solution aimed at refining the pruning metric formulation process. This initiative is geared towards enhancing the efficiency and efficacy of pruning within large language models.

\section{Detailed Search Space}

\subsection{Search Space Composition} 

Our search space comprises three specific input types: activations (X), gradients (G), and weights (W). These inputs provide complementary perspectives on the significance and contribution of different model components to the overall performance. 

\textbf{Activation.} Activation values from a model's layers offer rich, interpretable information about the role of individual neurons in processing input data. By considering activation magnitudes or patterns across a calibration dataset, we can identify and prune less important neurons or connections. For instance, consistently low activations might suggest redundancy, while highly correlated activations could point to opportunities for merging similar neurons.

\textbf{Gradient.} Gradients reflect the sensitivity of a model's output with respect to its parameters (weights). Larger gradient magnitudes indicate that small changes in the corresponding weights would significantly impact the model's predictions. Therefore, gradients can guide pruning by highlighting weights crucial for the model's performance. Pruning strategies might target weights with consistently small gradients, as their impact on the output is likely less significant.

\textbf{Weight.} Weights represent the learned parameters of a neural network and are fundamental to its ability to make predictions. The importance of weights can be assessed by evaluating their magnitudes and the roles they play within the network. Weights with larger magnitudes typically have a greater influence on the network's output, while smaller weights might contribute less significantly. By analyzing the distribution and significance of weights, we can identify which ones are essential for maintaining model performance. Pruning methods often focus on removing weights that have minimal impact, thereby simplifying the network and reducing its complexity without substantially degrading its accuracy.

\textbf{Hessian Matrix}, a fundamental concept in multivariable calculus, is instrumental in understanding the curvature of scalar fields and optimizing complex functions. While the utility of the Hessian matrix in various mathematical and engineering disciplines is well-documented, its specific application in the pruning of large language models (LLMs) merits particular attention due to its potential for enhancing model efficiency. The Hessian is a square matrix consisting of second-order partial derivatives of a scalar-valued function. For a function \( f(x_1, x_2, \ldots, x_n) \) of \( n \) variables, the Hessian matrix \( H \) is given by:

\[
H = \begin{bmatrix}
\frac{\partial^2 f}{\partial x_1^2} & \frac{\partial^2 f}{\partial x_1 \partial x_2} & \cdots & \frac{\partial^2 f}{\partial x_1 \partial x_n} \\
\frac{\partial^2 f}{\partial x_2 \partial x_1} & \frac{\partial^2 f}{\partial x_2^2} & \cdots & \frac{\partial^2 f}{\partial x_2 \partial x_n} \\
\vdots & \vdots & \ddots & \vdots \\
\frac{\partial^2 f}{\partial x_n \partial x_1} & \frac{\partial^2 f}{\partial x_n \partial x_2} & \cdots & \frac{\partial^2 f}{\partial x_n^2}
\end{bmatrix}
\]

This matrix is crucial for determining the local curvature of a function, which in turn helps in identifying maxima, minima, and saddle points — essential for optimization algorithms.

In the context of LLM pruning, the Hessian matrix offers a theoretically sound approach to evaluating the importance of weights within a neural network. Pruning methods aim to reduce the computational complexity of LLMs by removing weights that contribute minimally to the model's output. By assessing the impact of weight removal through the lens of the Hessian's curvature effects, one can theoretically predict and mitigate potential losses in model performance more effectively.

However, the practical application of the Hessian matrix in LLM pruning is not without challenges. The computation of the Hessian matrix is notably resource-intensive, particularly due to the necessity of calculating second-order derivatives for large matrices and potentially inverting these matrices to analyze their characteristics. These operations entail significant computational costs and can be prohibitive in terms of time, especially for models as large as those used in contemporary LLMs.

Given these considerations, our investigation into Hessian-based methods for LLM pruning not only enhances our theoretical understanding but also challenges us to develop more computationally feasible approaches. Future work could explore approximate methods to compute the Hessian or its relevant characteristics, such as using randomized algorithms or leveraging sparsity within the matrix to reduce the computational burden. By refining these methods, we aim to make Hessian-based pruning a practical tool for optimizing large-scale neural networks, ultimately contributing to more efficient and effective deployments of LLMs in real-world applications.

\subsection{Operation Vocabulary}\label{sec:operation_vocabulary}

As presented in Table~\ref{appendix:operation_vocabulary}, we provide operations vocabulary for describing common unary and binary functions used in mathematical and machine learning operations. The vocabulary is organized into two sections for unary and binary operations respectively. For the unary operations section, this Table lists 13 operations denoted by unique operation IDs from U01 to U13. For each operation, it provides a code example, a description of input and output addresses/types, and a mathematical definition of the operation. Some common unary functions included are square, negative value, absolute value, logarithm, exponent, square root, and tanh. The binary operations section similarly lists four common operations - addition, subtraction, multiplication, and division. These binary operations take two matrix inputs and produce a matrix output. Overall, this operations vocabulary provides a standardized indexing and definition of commonly used mathematical and machine learning building blocks to facilitate the description and replication of computational models and algorithms.

\subsection{Searched Metrics with Equations.} \label{appendix:searched_pruning_metric}

In the pursuit of optimizing neural network structures, Pruner-Zero has identified a set of novel pruning metrics, as illustrated in Table~\ref{appendix:tab:equation_searched}. These metrics are critical in the context of model simplification while maintaining, and in some instances, enhancing the performance of the neural network models. Each equation presented is accompanied by its Perplexity (PPL) score, which serves as an indicator of the model's predictive performance post-pruning. The scores close to the lower bound of this range suggest that the corresponding pruning metric has led to a model that balances the complexity and predictive capability effectively.

\begin{table}[t]
\centering
\caption{Pruning Metrics Discovered by Pruner-Zero with Corresponding Perplexity (PPL) Scores}\label{appendix:tab:equation_searched}
\vskip 0.15in
\begin{tabular}{cc}
\toprule
\textbf{Equation} & \textbf{PPL} \\
\midrule
$( \sigma\left(\left(|W| + |W|\right) + \sqrt{|G|}\right) )$ & 6.9175 \\
$( \log\left(\left|\left(|W| \times |W|\right)\right| + \left(G \times |W|\right)\right) )$ & 6.7483 \\
$( \left(\left(|W| + |W|\right) + \sqrt{|G|}\right) + \left(\frac{G}{\sqrt{\sqrt{|W|}}}\right) )$ & 6.9182 \\
$( (|W| + |W|) + \sqrt{|G|} )$ & 6.9175 \\
$( \tanh\left(\left||W| \times |W|\right|\right) + \left(\frac{G}{\sqrt{|W|}}\right) )$ & 6.9814 \\
$( \left|\left(|W| \times |W|\right)\right| \times \sqrt{|G|} )$ & 6.7701 \\
$( \sigma\left(\left|W\right|\right) \times \sigma(|W|) \times \sigma(|G|) )$ & 6.7079 \\
\bottomrule
\end{tabular}
\end{table}

\subsection{More Discussion over the Discovered SPM}\label{appendix:discuss}

We provide a concise theoretical analysis below and shed light on the insights gained from the investigated metric. Given an L-layer Large Language Model (LLM) denoted as \( W = \{w_0, \ldots, w_L\} \) and a dataset \( \mathcal{D} = \{(x_0, y_0), \ldots, (x_k, y_k)\} \), the task is to minimize the error \( E \) by optimizing the model parameters: 

\begin{equation}
\min_{W} E(\mathcal{D}, W) = \min_{W} E(y | x, W)
\end{equation}

In the context of pruning, the importance of a parameter is quantified by the error induced when the parameter is removed. The induced error can be measured as the squared difference with respect to a specific parameter \( w_m \):

\begin{equation}
I_m = \left( E(\mathcal{D}, W) - E(\mathcal{D}, W_{w_m = 0}) \right)^2
\end{equation}

To approximate this error, we can use a second-order Taylor expansion:

\begin{equation}
I_m^{(2)}(W) = \left( g_m w_m - \frac{1}{2} w_m H_m W \right)^2
\end{equation}

where \( g_m \) is the gradient and \( H_m \) is the Hessian matrix. However, computing the Hessian matrix is computationally expensive, thus we employ a first-order expansion to obtain a more tractable approximation:

\begin{equation}
I_m^{(1)}(W) = \left( g_m w_m \right)^2
\end{equation}

This can be further simplified to:

\begin{equation}
I(W) = (W \times G)^2
\end{equation}

where \( W \) is the weight vector and \( G \) is the gradient vector.

\textbf{Insight 1: Measurement of Kurtosis.} We measure the kurtosis of the weights and gradients for the LLaMA-2-7B model, obtaining values of 0.7734 and 5.8203, respectively. This indicates that both weights and gradients significantly contribute to the Sparsity Promoting Metric (SPM) and influence the relative importance of different weights. To balance their contributions, we employed an evolutionary algorithm to automatically find the optimal scaling factors.

\textbf{Insight 2: Inspired by Quantization Methods.} Inspired by quantization methods such as AWQ~\cite{lin2023awq}, SmoothQuant~\cite{xiao2023smoothquant}, and LLM-FP4~\cite{liu-etal-2023-llm-fp4}, we employ scaling factors to balance weights and activations. This step is critical in reducing computational requirements without significantly compromising model accuracy. This approach underscores the importance of balancing weights and gradients when evaluating weight significance.

Based on these insights, we simplify Equation~\ref{eq:searched_spm} to:

\begin{equation}
I(W) = (s_1(W) \times s_2(G))^2
\end{equation}

Here, \( s_1 \) and \( s_2 \) are scaling functions that control the contributions of weights and gradients to the final importance metric. Our goal is to identify these functions. Specifically, \( s_1 \) is a square operation that penalizes minor weights, and \( s_2 \) is a mean-median scaling (mms) operation that moderates the influence of gradients, preventing excessive dominance due to scaling disparities.

In Appendix~\ref{sec:operation_vocabulary}, we analyze the impact of these operations on final performance by using a correlation matrix to identify the most influential operations. Our findings highlight the critical role of the mms operation in optimizing the balance between weight and gradient contributions. 

By systematically investigating and refining these scaling operations, we achieve a more nuanced understanding of parameter importance in LLM pruning. This approach not only enhances computational efficiency but also preserves model performance, demonstrating the efficacy of our proposed metric in practical applications.

\section{Genetic Programming in SPM Optimization}

In our study, we utilize genetic programming as outlined in Algorithm~\ref{alg:evolution} to optimize the search space for pruning metrics in large language models. Genetic programming is a type of evolutionary algorithm that automates the generation of computer programs to solve specific computational problems. Our application of this methodology focuses on the development of symbolic pruning metrics for neural networks.

\textbf{Implementation Details:}
Each individual in the genetic programming population represents a potential pruning metric, where the structure of each individual is modeled as a symbolic tree. Nodes within these trees correspond to either unary or binary operations, crucial for defining the computational logic of the pruning metric.

\textbf{Evolutionary Process:}
The evolutionary process unfolds through several distinct stages, ensuring the continuous improvement of the population with respect to the pruning task:

\begin{enumerate}
    \item \textbf{Initialization:}
    The process begins with the generation of a random population of symbolic trees. This initial population forms the basis from which better-suited solutions can evolve over successive generations.

    \item \textbf{Selection:}
    We implement a selection strategy where the top-k symbolic pruning metrics are chosen based on their performance. From this subset, two candidates are randomly sampled to serve as parents for the next generation. This selective approach ensures that only the most promising metrics contribute to the gene pool.

    \item \textbf{Crossover:}
    The crossover operation is vital for introducing variability and combining beneficial traits from two parents. It involves scanning each parent using Depth-First Search (DFS) to identify a node. A node from each parent is then randomly selected, and their subtrees are exchanged to produce offspring. This method of recombination allows offspring to inherit and reconfigure traits from both parents, potentially leading to more effective pruning metrics.

    \item \textbf{Mutation:}
    After crossover, each node in the offspring undergoes a mutation with a probability of 50\%. This mutation can alter the node's operation, encouraging diversity within the population and aiding in the exploration of the search space.
\end{enumerate}

\begin{figure}[t]
    \centering
    \includegraphics[width=0.7\linewidth]{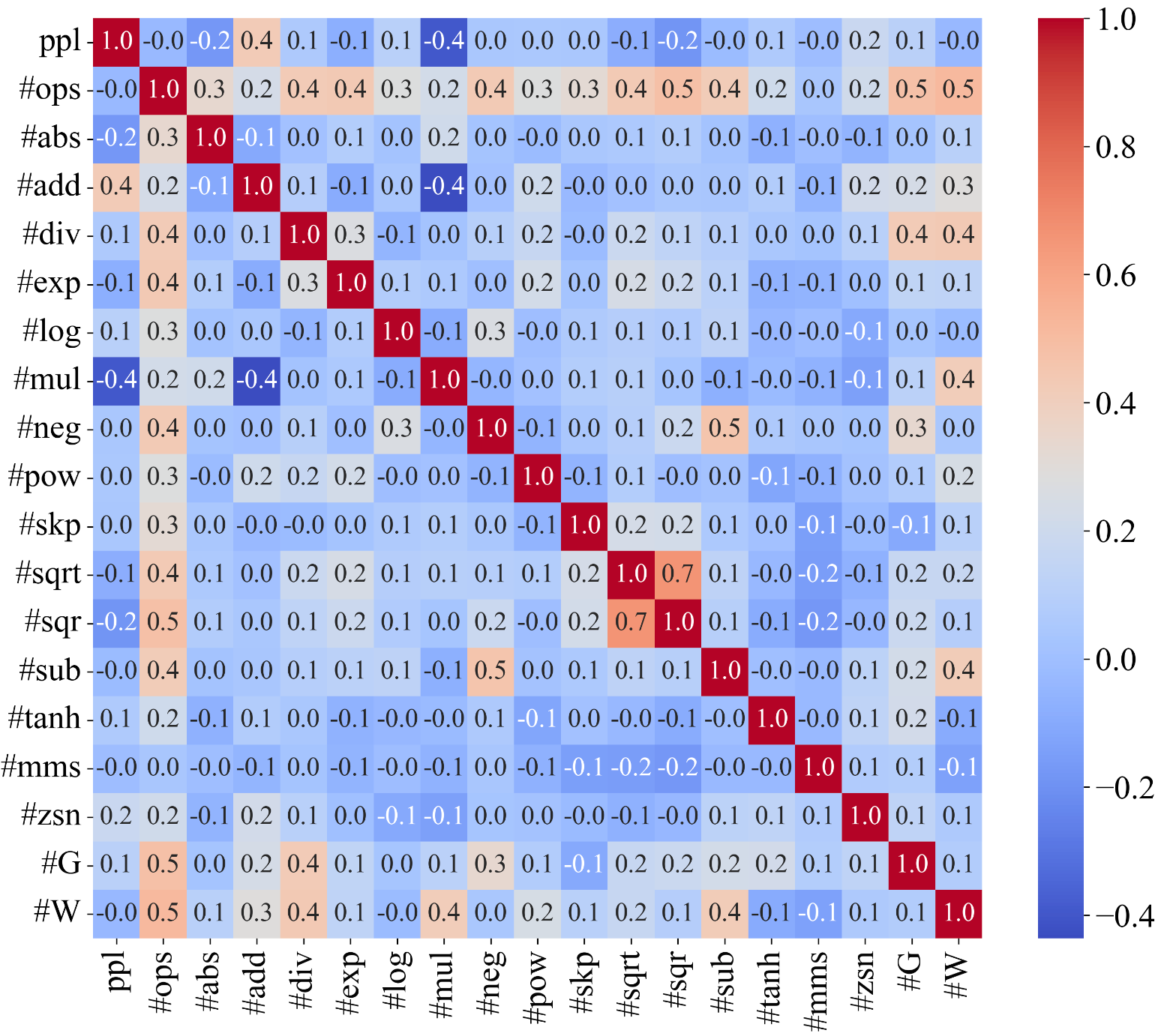}
    \caption{Correlation Matrix of Primitive Operations with Perplexity.}
    \label{fig:correlation_between_operations}
\end{figure}

\textbf{Evaluation of Offspring:}
The newly generated offspring, now representing different symbolic pruning metrics, are evaluated based on their fitness. Fitness is measured by the perplexity of the pruned model, specifically the LLaMA-2-7B model under 50\% sparsity. This measure helps determine how well the pruning metric performs in reducing the model's complexity without significantly impacting its effectiveness.

\textbf{Iterative Process:}
The selection, crossover, and mutation processes are repeated for \(\mathcal{N}\) iterations, or until a satisfactory pruning metric is found. Each iteration refines the population, ideally leading to increasingly effective pruning strategies over time.

By employing genetic programming, we leverage a robust evolutionary framework to explore a diverse array of pruning metrics systematically. This approach not only aids in discovering highly effective pruning techniques but also contributes to the broader understanding of how symbolic operations can impact the performance of large neural networks under sparsity constraints.

\section{Expanding the Results}

\subsection{Detailed Correlation Analysis}
\label{appendix:detailed_correlation_analysis}

\textbf{Methodology.} During the evolutionary search phase, a comprehensive dataset of symbolic pruning metrics was established alongside their associated perplexity scores. Metrics yielding a perplexity score below 7 were earmarked as potential candidates for effective pruning strategies. These metrics are systematically enumerated in Table~\ref{appendix:tab:equation_searched} within Appendix~\ref{appendix:searched_pruning_metric}.

\textbf{Visualization and Interpretation.} Figure~\ref{fig:correlation_between_operations} visualizes the correlation coefficients between the frequency of various operations within the potential metrics and their corresponding perplexity scores. A heatmap representation facilitates an intuitive understanding of these relationships. The following points summarize the major findings from this analysis:

\begin{itemize}
    \item Multiplication Operations (\#mul): A strong positive correlation with perplexity was observed for multiplication operations, denoted by \#mul in the heatmap. The frequency of multiplication within a metric tends to coincide with lower perplexity scores, suggesting that incorporating multiplication may be beneficial for pruning effectiveness.
    \item Opposite Operations (\#sqrt and \#sqr): A noteworthy correlation was detected between square roots (\#sqrt) and squaring (\#sqr) operations, indicating that they often occur in tandem and may have compensatory dynamics within the metrics.
    \item Min-Max Scaling (\#mms): The min-max scaling operation, referred to as \#mms in the heatmap, showed minimal correlation with perplexity and other operations. This implies that mms contributes a stabilizing effect on the metric's performance, reinforcing its utility in the pruning process.
\end{itemize}

\textbf{Conclusion and Implications for Metric Design.} The insights gleaned from this correlation analysis provide a data-driven foundation for refining symbolic pruning metrics. Specifically, the importance of multiplication and the potential balancing act between opposite operations offer avenues for enhancing metric sophistication. Meanwhile, the robust nature of min-max scaling underscores its value as a consistent component in pruning metrics. These analytical observations have been instrumental in the formulation of the symbolic pruning metrics presented in Section~\ref{genetic_programming_framework}, and may inform future developments in pruning methodology.

\begin{table}[t]
\centering
\caption{Data from trials at various depths}
\vskip 0.15in

\label{tab:redundency_in_ss}
\begin{tabular}{ccccccc}
\toprule
\textbf{Depth} & \textbf{Trial1} & \textbf{Trial2} & \textbf{Trial3} & \textbf{Trial4} & \textbf{Average} & \textbf{Standard deviation} \\ \midrule
3     & 0.34   & 0.25   & 0.33   & 0.32   & 0.31    & 0.035    \\ 
5     & 0.33   & 0.34   & 0.26   & 0.24   & 0.29    & 0.043    \\ 
7     & 0.27   & 0.26   & 0.15   & 0.24   & 0.23    & 0.047    \\ 
9     & 0.26   & 0.31   & 0.31   & 0.26   & 0.28    & 0.027    \\ \bottomrule
\end{tabular}
\end{table}

\begin{table*}[t]
\centering
\caption{Perplexity of pruned LLaMA models on WikiText2. Our \method{} outperforms SparseGPT~\cite{Frantar2023SparseGPTML}, Wanda~\cite{Sun2023ASA_wanda} and GBLM-Pruner~\cite{Das2023BeyondSH_GBLM}, achieving lower Perplexity without weight updates.}
\label{appendix:tab:gblm-pruner}
\vskip 0.15in

\begin{adjustbox}{max width=\textwidth}
\begin{tabular}{lcccc}
\toprule
\textbf{Method}        & \textbf{Sparsity} & \textbf{LLaMA-7B} & \textbf{LLaMA-13B} & \textbf{LLaMA-30B} \\ \midrule
Dense          & 0        & 5.68     & 5.09      & 4.10       \\ \hline
Magnitude~\cite{song2016deep_compression}     & 0.5      & 17.29    & 20.21     & 7.54       \\
SparseGPT~\cite{Frantar2023SparseGPTML}     & 0.5      & 7.22     & 6.19      & 5.32       \\
Wanda~\cite{Sun2023ASA_wanda}         & 0.5      & 7.26     & 6.15      & 5.24       \\
GBLM-Pruner2~\cite{Das2023BeyondSH_GBLM}  & 0.5      & 7.19     & 6.14      & 5.23       \\
GBLM-Pruner1~\cite{Das2023BeyondSH_GBLM}  & 0.5      & 7.15     & 6.11      & 5.18       \\ 
\method(Ours) & 0.5      & \textbf{6.95} & \textbf{5.94} & \textbf{5.01} \\ \hline
Magnitude~\cite{song2016deep_compression}     & 2:4      & 42.54    & 18.36     & 9.11       \\
SparseGPT~\cite{Frantar2023SparseGPTML}     & 2:4      & 10.88    & 9.0       & 7.12       \\
Wanda~\cite{Sun2023ASA_wanda}         & 2:4      & 11.53    & 9.59      & 6.90       \\
GBLM-Pruner2~\cite{Das2023BeyondSH_GBLM}  & 2:4      & 11.36    & 9.45      & 6.90       \\
GBLM-Pruner1~\cite{Das2023BeyondSH_GBLM}  & 2:4      & 11.33    & 9.16      & 6.87       \\ 
\method(Ours) & 2:4      & \textbf{10.61} & \textbf{8.11} & \textbf{6.51} \\ \hline
Magnitude~\cite{song2016deep_compression}     & 4:8      & 16.83    & 13.87     & 7.62       \\
SparseGPT~\cite{Frantar2023SparseGPTML}     & 4:8      & 8.45     & 7.44      & 6.18       \\
Wanda~\cite{Sun2023ASA_wanda}         & 4:8      & 8.57     & 7.41      & 5.97       \\
GBLM-Pruner2~\cite{Das2023BeyondSH_GBLM}  & 4:8      & 8.50     & 7.38      & 5.94       \\
GBLM-Pruner1~\cite{Das2023BeyondSH_GBLM}  & 4:8      & 8.48     & 7.26      & 5.89       \\ 
\method(Ours) & 4:8      & \textbf{8.12} & \textbf{6.81} & \textbf{5.65} \\ \bottomrule
\end{tabular}
\end{adjustbox}
\end{table*}

\subsection{Expanding the Zero-Shot Tasks}\label{appendix:zero_shot}

\textbf{Detailed Results of Zero-shot Tasks.} In the context of zero-shot learning, our evaluation encompasses a diverse set of tasks, as presented in Table~\ref{tab:zero_shot_results}. These tasks include BoolQ~\citep{Clark2019BoolQET}, RTE~\citep{Wang2018GLUEAM}, HellaSwag~\citep{Zellers2019HellaSwagCA}, WinoGrande~\citep{Sakaguchi2019WinoGrande}, ARC Easy and Challenge~\citep{Clark2018ThinkYH}, and OpenbookQA~\citep{Mihaylov2018CanAS}. To ensure the reproducibility of our results, we adhere to the settings and methodologies outlined in the Wanda study~\cite{Sun2023ASA_wanda}. Detailed task-wise performance metrics are systematically presented across several tables: Table~\ref{appendix:tab:zs_llama_unstructured} demonstrates the unstructured evaluation, while Tables~\ref{appendix:tab:zs_llama_48} and \ref{appendix:tab:zs_llama_24} provide insights into structured evaluations. Furthermore, the nuanced performances under the LLaMA framework are captured in Tables~\ref{appendix:tab:zs_LLaMA-2_unstructured}, \ref{appendix:tab:zs_LLaMA-2_48}, and \ref{appendix:tab:zs_LLaMA-2_24}, offering a comprehensive understanding of the zero-shot capabilities of the models in question.

\textbf{Discussion.} Pruning involves removing redundant or less important parameters from the model, which can act as a form of regularization or denoising. (1) Regularization: Pruning can act as a form of regularization~\cite{Wang2020NeuralPV}, reducing overfitting by applying dropout to the model. Dense models might overfit the training data or learn spurious correlations that do not generalize well to unseen tasks. In contrast, the pruned model might generalize better to new tasks due to this implicit regularization. (2) Denoising: In dense models, some parameters act as noise~\cite{diao2023pruning}, contributing little to the model's performance or even detracting from it. Pruning eliminates much of this noise and thus leads to a clearer signal within the model. This denoising effect can make weakly learned facts or knowledge more accessible to the model, leading to improved performance on zero-shot tasks. For instance, LASER~\cite{sharma2024the} finds that even under extreme reductions, the performance of LLM on natural language understanding tasks continues to improve.
Recent research~\cite{Li2024EvaluatingQL} suggests a similar phenomenon in quantization, observing that quantization even brings notable accuracy gain for multi-choice zero-shot tasks. They conducted experiments to verify that there is less uncertainty for the quantized model, resulting in the phenomenon that quantization brings accuracy gain. Additionally, previous methods like Wanda~\cite{Sun2023ASA_wanda} and SparseGPT~\cite{Frantar2023SparseGPTML} have also demonstrated improved performance on zero-shot tasks after pruning. This suggests that these post-training pruning techniques can be effective in enhancing the reasoning capabilities of language models, particularly in scenarios where they need to rely on their pre-trained knowledge without task-specific fine-tuning.

\begin{table*}[bt]
  \centering
  \caption{
  Accuracies ($\%$) of LLaMA for 7 zero-shot tasks with unstructured 50$\%$ sparsity.
  }
  \vskip 0.15in
\resizebox{.85\textwidth}{!}{
  \setlength{\tabcolsep}{5.5pt}
  
  \begin{tabular}{crccccccccc}
 \midrule 
\textbf{Params}  & \hspace{-0.4cm} \textbf{Method} & \hspace{-0.2cm} \textbf{BoolQ} & \textbf{RTE} & \hspace{-0.35cm} \textbf{HellaSwag}  & \hspace{-0.3cm} \textbf{WinoGrande} & \hspace{-0.2cm} \textbf{ARC-e} & \textbf{ARC-c} & \textbf{OBQA} & \textbf{Mean} \\
\midrule 
  \multirow{5}{*}{7B}   & Dense    & 75.05 & 66.43 & 56.92 & 69.93 & 75.34 & 41.89 & 34.40 & 59.99  \\
  \cmidrule{2-10}
  & Magnitude & 54.59 & 54.51 & 45.49 & 59.19 & 58.84 & 33.53 & 22.40 & 46.94\\
  & SparseGPT & \tbf{72.05} & 54.15 & 51.43 & \tbf{67.88} & \tbf{71.38} & 37.71 & 30.00 & 54.94\\ 
  &   Wanda & 71.22 & 55.60 & 51.85 & 66.06 & 69.11 & 36.86 & 28.80 & 54.21 \\ 
  \gr \wc   &   \method{}& 68.87 & \textbf{63.54} & \textbf{69.12} & 67.80 & 69.19 & \textbf{38.82} & \textbf{39.60} & \textbf{59.56} \\ 
\midrule      
  \multirow{4}{*}{13B}   & Dense    &  77.89 & 70.4 & 59.94 & 72.77 & 77.40 & 46.50 & 33.20 & 62.59  \\
  \cmidrule{2-10}
  & Magnitude & 54.89 & 51.26 & 44.16 & 63.14 & 58.80 & 33.79 & 27.20 & 47.61 \\
  & SparseGPT & \tbf{76.97} & 61.01 & 54.95 & 71.67 & 72.47 & 41.98 & 31.20 & 58.61 \\ 
  & Wanda & 75.90 & \tbf{62.82} & 55.71 & \tbf{71.98} & 73.19 & \tbf{43.52} & 32.20 & 59.33 \\ 
 \gr \wc  &  \method{}& 74.77 & 59.57 & \textbf{74.00} & 71.19 & \textbf{74.12} & 42.06 & \textbf{43.00} & \textbf{62.67} \\ 
\midrule     
  \multirow{4}{*}{30B}   & Dense & 82.69 & 66.79 & 63.35 & 75.69 & 80.30 & 52.82 & 36.00 & 65.38\\
  \cmidrule{2-10}
  & Magnitude & 64.34 & 50.18 & 50.59 & 66.54 & 72.39 & 43.77 & 29.00 & 53.83 \\
  & SparseGPT & \tbf{82.32} & 62.45 & 59.15 & \tbf{75.22} & 78.96 & 48.56 & 35.00 & 63.09   \\ 
  & Wanda & 81.90 & \tbf{65.34} & 60.93 & 73.48 & \tbf{79.29} & \tbf{49.66} & 34.60 & 63.60 \\ 
\gr \wc   &  \method{}& 81.04 & 67.15 & \textbf{79.25} & 73.32 & 78.07 & 48.81 & \textbf{44.80} & \textbf{67.49} \\ 
\midrule     
  \multirow{4}{*}{65B}   & Dense    & 84.83 & 69.68 & 64.54 & 77.27 & 81.40 & 52.90 & 38.20 & 66.97 \\
  \cmidrule{2-10}
  & Magnitude & 79.15 & 62.45 & 61.90 & 74.74 & 76.40 & 49.57 & 35.00 & 62.74 \\
  & SparseGPT & 84.60 & 70.76 & 63.90 & \tbf{77.43} & 79.35 & 50.85 & 37.20 & 66.30 \\ 
  & Wanda & \tbf{84.70} & \tbf{71.48} & 64.55 & 76.87 & \tbf{79.75} & 50.51 & 38.80 & 66.67\\
\gr \wc  &  \method{}& 84.45 & 70.04 & \textbf{81.75} & 75.77 & 79.45 & \textbf{52.65} & \textbf{44.60} & \textbf{69.81}\\
  \bottomrule 
  \end{tabular}
  }
  \label{appendix:tab:zs_llama_unstructured}
\end{table*}

\begin{table*}[bt]
  \centering
  \caption{
  Accuracies ($\%$) of LLaMA for 7 zero-shot tasks with 4:8 sparsity. 
  }
  \vskip 0.15in
\resizebox{.85\textwidth}{!}{
  \setlength{\tabcolsep}{5.5pt}
  \begin{tabular}{crccccccccc}
  \toprule
\textbf{Params}  & \hspace{-0.4cm} \textbf{Method} & \hspace{-0.2cm} \textbf{BoolQ} & \textbf{RTE} & \hspace{-0.35cm} \textbf{HellaSwag}  & \hspace{-0.3cm} \textbf{WinoGrande} & \hspace{-0.2cm} \textbf{ARC-e} & \textbf{ARC-c} & \textbf{OBQA} & \textbf{Mean} \\
  \midrule
  \multirow{4}{*}{7B}   & Dense    &  75.05 & 66.43 & 56.92 & 69.93 & 75.34 & 41.89 & 34.40 & 59.99 \\
  \cmidrule{2-10}
  & Magnitude & 51.19 & 50.54 & 46.73 & 60.69 & 58.96 & 30.89 & 23.20 & 46.03 \\
  & SparseGPT & \tbf{73.06} & 58.12 & 47.88 & \tbf{65.98} & \tbf{66.75} & 32.42 & 25.40 & 52.80  \\ 
  & Wanda & 70.97 & 58.24 & 46.81 & 65.83 & 65.53 & 33.97 & 28.00 & 52.76 \\ 
\gr \wc   &   \method{}& 69.97 & \textbf{60.65} & \textbf{63.70} & 64.88 & 64.23 & \textbf{34.64} & \textbf{35.60} & \textbf{56.24} \\ 
  \midrule   
  \multirow{4}{*}{13B}   & Dense    &  77.89 & 70.40 & 59.94 & 72.77 & 77.40 & 46.50 & 33.20 & 62.59 \\
  \cmidrule{2-10}
  & Magnitude & 61.07 & 51.26 & 48.91 & 65.11 & 63.26 & 35.67 & 28.40 & 50.53  \\
  & SparseGPT & \tbf{76.61} & 57.76 & 51.24 & 70.17 & \tbf{71.17} & 37.20 & 27.80 & 55.99 \\ 
  & Wanda & 74.89 & \tbf{57.89} & 51.26 & \tbf{70.56} & 70.29 & \tbf{37.97} & 29.80 & 56.09 \\ 
 \gr \wc  &  \method{}& 69.20 & 53.07 & \textbf{70.01} & 69.22 & 70.50 & 39.42 & \textbf{41.80} & \textbf{59.03} \\ 
  \midrule   
  \multirow{4}{*}{30B}   & Dense    &  82.69 & 66.79 & 63.35 & 75.69 & 80.30 & 52.82 & 36.00 & 65.38 \\
  \cmidrule{2-10}
  & Magnitude & 63.55 & 50.18 & 49.45 & 65.75 & 73.36 & 42.83 & 29.60 & 53.53 \\
  & SparseGPT & \tbf{78.69} & \tbf{61.73} & 56.15 & \tbf{74.35} & 76.94 & 46.08 & 31.60 & 60.79 \\ 
  & Wanda & 77.38 & 58.80 & 58.79 & 74.28 & \tbf{77.34} & \tbf{46.46} & 34.00 & 61.00 \\ 
\gr \wc   &  \method{}& 77.71 & 58.48 & \textbf{75.76} & 72.14 & 75.72 & 45.90 & \textbf{42.60} & \textbf{64.04} \\ 
  \midrule  
  \multirow{4}{*}{65B}   & Dense    &    84.83 & 69.68 & 64.54 & 77.27 & 81.40 & 52.90 & 38.20 & 66.97 \\
  \cmidrule{2-10}
  & Magnitude & 74.95 & 68.23 & 60.85 & 74.27 & 76.45 & 47.61 & 32.80 & 62.17 \\
  & SparseGPT & \tbf{84.35} & 68.95 & 61.00 & \tbf{77.19} & 78.75 & 48.46 & 35.40 & 64.87  \\ 
  & Wanda & 84.29 & \tbf{70.92} & 59.54 & 76.64 & \tbf{79.00} & 48.83 & 35.60 & 64.97 \\
\gr \wc  &  \method{}& 81.50 & 66.43 & \textbf{79.15} & 76.48 & 78.50 & \textbf{50.00} & \textbf{44.20} & \textbf{68.04} \\
  \bottomrule  
  \end{tabular}
  }

  \label{appendix:tab:zs_llama_48}
\end{table*}

\begin{table*}[bt]
  \centering

\caption{
  Accuracies ($\%$) of LLaMA for 7 zero-shot tasks with 2:4 sparsity.
  }
  \vskip 0.15in
\resizebox{.8\textwidth}{!}{
  \setlength{\tabcolsep}{5.5pt}
  \begin{tabular}{crccccccccc}
  \toprule
\textbf{Params}  & \hspace{-0.4cm} \textbf{Method} & \hspace{-0.2cm} \textbf{BoolQ} & \textbf{RTE} & \hspace{-0.35cm} \textbf{HellaSwag}  & \hspace{-0.3cm} \textbf{WinoGrande} & \hspace{-0.2cm} \textbf{ARC-e} & \textbf{ARC-c} & \textbf{OBQA} & \textbf{Mean} \\
  \midrule
  \multirow{4}{*}{7B}   & Dense    &  75.05 & 66.43 & 56.92 & 69.93 & 75.34 & 41.89 & 34.40 & 59.99 \\
  \cmidrule{2-10}
  & Magnitude &  53.09 & 55.60 & 42.30 & 59.91 & 53.28 & 27.13 & 21.80 & 44.73 \\
  & SparseGPT & \tbf{70.46} & \tbf{60.65} & 42.99 & \tbf{64.88} & \tbf{61.49} & 30.12 & 23.60 & 50.60 \\ 
  & Wanda & 69.30 & 51.99 & 42.06 & 62.75 & 60.94 & 28.07 & 24.60 & 48.53 \\ 
\gr \wc   &   \method{}& 66.54 & 53.79 & \textbf{56.29} & 62.19 & 60.56 & 30.63 & 34.40 & \textbf{52.06} \\ 
  \midrule   
  \multirow{4}{*}{13B}   & Dense    &   77.89 & 70.40 & 59.94 & 72.77 & 77.40 & 46.50 & 33.20 & 62.59 \\
  \cmidrule{2-10}
  & Magnitude & 60.95 & 49.10 & 45.81 & 62.75 & 58.75 & 31.06 & 27.60 & 48.00 \\
  & SparseGPT & \tbf{72.14} & \tbf{55.23} & 48.11 & \tbf{68.98} & 66.71 & 34.98 & 26.40 & 53.22 \\ 
  & Wanda & 70.21 & 53.43 & 46.74 & 68.82 & 65.82 & 33.87 & 27.20 & 52.30 \\ 
 \gr \wc  &  \method{}& 69.20 & 53.07 & \textbf{64.67} & 68.51 & \textbf{67.13} & \textbf{36.09} & \textbf{38.80} & \textbf{56.78} \\ 
  \midrule   
  \multirow{4}{*}{30B}   & Dense &   82.69 & 66.79 & 63.35 & 75.69 & 80.30 & 52.82 & 36.00 & 65.38 \\
  \cmidrule{2-10}
  & Magnitude & 65.11 & 52.35 & 51.72 & 66.22 & 70.88 & 38.23 & 27.60 & 53.16  \\
  & SparseGPT & \tbf{75.60} & 62.13 & 53.10 & 72.61 & \tbf{75.13} & 41.98 & 31.80 & 58.91 \\ 
  & Wanda & 74.68 & \tbf{63.80} & 54.41 & \tbf{72.93} & 74.41 & 42.06 & 32.20 & 59.21 \\ 
\gr \wc   &  \method{}& 70.92 & 62.09 & \textbf{72.47} & 70.24 & 73.36 & \textbf{42.32} & \textbf{42.60} & \textbf{62.00} \\ 
  \midrule   
  \multirow{4}{*}{65B}   & Dense    &   84.83 & 69.68 & 64.54 & 77.27 & 81.40 & 52.90 & 38.20 & 66.97 \\
  \cmidrule{2-10}
  & Magnitude & 77.9 & 64.98 & 58.65 & 72.85 & 75.15 & 45.05 & 34.40 & 61.28 \\
  & SparseGPT & 83.15 & 65.34 & 57.20 & \tbf{76.72} & 78.20 & 45.18 & 32.20 & 62.57 \\ 
  & Wanda & \tbf{83.58} & \tbf{66.79} & 56.36 & 75.82 & \tbf{78.23} & 45.56 & 33.60 & 62.84 \\
\gr \wc  &  \method{}& 79.45 & 60.29 & \textbf{76.95} & 73.48 & 77.20 & \textbf{46.76} & \textbf{43.80} & \textbf{65.42} \\
  \bottomrule 
  \end{tabular}
  }

  \label{appendix:tab:zs_llama_24}
\end{table*}

\begin{table}[bt]
  \centering

  \caption{
  Accuracies ($\%$) of LLaMA-2 for 7 zero-shot tasks with unstructured 50$\%$ sparsity.
  }
  \vskip 0.15in
\resizebox{.8\textwidth}{!}{
  \setlength{\tabcolsep}{5.5pt}
  \begin{tabular}{crccccccccc}
  \toprule
\textbf{Params}  & \hspace{-0.4cm} \textbf{Method} & \hspace{-0.2cm} \textbf{BoolQ} & \textbf{RTE} & \hspace{-0.35cm} \textbf{HellaSwag}  & \hspace{-0.3cm} \textbf{WinoGrande} & \hspace{-0.2cm} \textbf{ARC-e} & \textbf{ARC-c} & \textbf{OBQA} & \textbf{Mean} \\
  \midrule
  \multirow{4}{*}{7B}   & Dense    &  77.74 & 62.82 & 57.17 & 68.90 & 76.39 & 43.52 & 31.40 & 59.71\\
  \cmidrule{2-10}
  & Magnitude & 63.00 & \tbf{57.04} & 49.13 & 63.30 & 64.10 & 34.64 & 26.80 & 51.14 \\
  & SparseGPT & 75.02 & 54.15 & 52.37 & \tbf{69.85} & \tbf{73.27} & \tbf{39.85} & 29.20 & 56.24 \\ 
  & Wanda & \tbf{75.99} & 53.43 & 52.49 & 68.19 & 72.77 & 39.59 & 31.20 & 56.24 \\ 
\gr \wc   &   \method{}& 70.92 & 53.07 & \textbf{69.10} & 65.98 & 71.89 & 39.76 & \textbf{41.40} & \textbf{58.87} \\ 
  \midrule   
  \multirow{4}{*}{13B}   & Dense    &  80.52 & 65.34 & 60.06 & 72.22 & 79.42 & 48.46 & 35.20 & 63.03 \\
  \cmidrule{2-10}
  & Magnitude & 57.61 & 55.96 & 54.40 & 65.27 & 70.54 & 38.40 & 27.80 & 52.85 \\
  & SparseGPT & 81.44 & 65.34 & 55.83 & \tbf{72.77} & 74.83 & 42.24 & 32.60 & 60.72 \\ 
  & Wanda & \tbf{81.84} & 64.02 & 56.90 & 71.35 & \tbf{76.18} & 43.52 & 32.00 & 60.83 \\ 
 \gr \wc  &  \method{}& 79.79 & \textbf{66.06} & \textbf{74.60} & 70.88 & 75.67 & \textbf{43.60} & \textbf{43.20} & \textbf{64.83} \\ 
  \midrule   
  \multirow{4}{*}{70B}   & Dense    & 83.40 & 67.87 & 66.10 & 78.06 & 82.55 & 54.44 & 37.20 & 67.08 \\
  \cmidrule{2-10}
  & Magnitude & 70.55 & 60.65 & 61.50 & 73.48 & 75.70 & 49.23 & 35.40 & 60.93 \\
  & SparseGPT & \tbf{83.55} & 70.40 & 63.80 & \tbf{78.85} & \tbf{82.40} & 53.75 & 38.20 & 67.28 \\ 
  & Wanda & 82.50 & \tbf{73.65} & 64.10 & 78.14 & 80.80 & 52.65 & 37.40 & 67.03 \\
\gr \wc  &  \method{}& 83.20 & 73.29 & \textbf{81.70} & 76.72 & 81.10 & \textbf{55.12} & \textbf{46.60} & \textbf{71.10} \\
  \bottomrule 
  \end{tabular}
  }
  \label{appendix:tab:zs_LLaMA-2_unstructured}
\end{table}

\begin{table*}[bt]
  \centering

\caption{
  Accuracies ($\%$) of LLaMA-2 for 7 zero-shot tasks with 4:8 sparsity.
  }
  \vskip 0.15in
\resizebox{.8\textwidth}{!}{
  \setlength{\tabcolsep}{5.5pt}
  \begin{tabular}{crccccccccc}
  \toprule
\textbf{Params}  & \hspace{-0.4cm} \textbf{Method} & \hspace{-0.2cm} \textbf{BoolQ} & \textbf{RTE} & \hspace{-0.35cm} \textbf{HellaSwag}  & \hspace{-0.3cm} \textbf{WinoGrande} & \hspace{-0.2cm} \textbf{ARC-e} & \textbf{ARC-c} & \textbf{OBQA} & \textbf{Mean} \\
  \midrule
  \multirow{4}{*}{7B}   & Dense    &  77.74 & 62.82 & 57.17 & 68.90 & 76.39 & 43.52 & 31.40 & 59.71\\
  \cmidrule{2-10}
  & Magnitude & 63.00 & 52.35 & 50.08 & 62.43 & 64.73 & 35.92 & 26.00 & 50.64 \\
  & SparseGPT & 72.69 & \tbf{55.23} & 48.20 & \tbf{68.11} & \tbf{69.15} & 35.84 & 27.40 & 53.80 \\ 
  & Wanda & \tbf{73.91} & 53.79 & 46.45 & 66.61 & 66.71 & 34.13 & 25.80 & 52.49 \\ 
\gr \wc   &   \method{}& 67.52 & 53.43 & \textbf{63.06} & 64.48 & 66.96 & \textbf{36.69} & \textbf{38.60} & \textbf{55.82} \\ 
  \midrule   
  \multirow{4}{*}{13B}   & Dense    &  80.52 & 65.34 & 60.06 & 72.22 & 79.42 & 48.46 & 35.20 & 63.03 \\
  \cmidrule{2-10}
  & Magnitude & 63.33 & 57.76 & 53.96 & 64.40 & 68.48 & 35.75 & 26.00 & 52.81 \\
  & SparseGPT & 79.97 & \tbf{66.79} & 52.01 & \tbf{70.64} & 73.61 & 41.04 & 30.00 & 59.15 \\ 
  & Wanda & \tbf{80.26} & 65.62 & 52.05 & 69.48 & 73.88 & 41.54 & 28.40 & 58.75\\ 
 \gr \wc  &  \method{}& 79.45 & 60.65 & \textbf{70.00} & 67.64 & \textbf{74.03} & \textbf{41.64} & \textbf{40.40} & \textbf{61.97}\\ 
  \midrule   
  \multirow{4}{*}{70B}   & Dense    &  83.40 & 67.87 & 66.10 & 78.06 & 82.55 & 54.44 & 37.20 & 67.08 \\
  \cmidrule{2-10}
  & Magnitude & 70.95 & 59.21 & 60.05 & 74.11 & 76.25 & 46.76 & 34.60 & 60.28 \\
  & SparseGPT & 82.20 & \tbf{72.20} & 61.45 & \tbf{77.82} & \tbf{80.85} & 51.19 & 35.20 & 65.84 \\ 
  & Wanda & \tbf{84.30} & 71.80 & 61.90 & 76.24 & 80.40 & 51.80 & 36.00 & 66.06 \\
\gr \wc  &  \method{}& 83.55 & 71.12 & \textbf{80.40} & 75.37 & 80.45 & \textbf{52.47} & \textbf{46.20} & \textbf{69.94} \\
  \bottomrule 
  \end{tabular}
  }
  \label{appendix:tab:zs_LLaMA-2_48}
\end{table*}

\begin{table*}[bt]
  \centering

    \caption{
  Accuracies ($\%$) of LLaMA-2 for 7 zero-shot tasks with 2:4 sparsity.
  }
  \vskip 0.15in
\resizebox{.85\textwidth}{!}{
  \setlength{\tabcolsep}{5.5pt}
  \begin{tabular}{crccccccccc}
  \toprule
\textbf{Params}  & \hspace{-0.4cm} \textbf{Method} & \hspace{-0.2cm} \textbf{BoolQ} & \textbf{RTE} & \hspace{-0.35cm} \textbf{HellaSwag}  & \hspace{-0.3cm} \textbf{WinoGrande} & \hspace{-0.2cm} \textbf{ARC-e} & \textbf{ARC-c} & \textbf{OBQA} & \textbf{Mean} \\
  \midrule
  \multirow{4}{*}{7B}   & Dense    & 77.74 & 62.82 & 57.17 & 68.90 & 76.39 & 43.52 & 31.40 & 59.71 \\
  \cmidrule{2-10}
  & Magnitude & 56.23 & 51.35 & 42.27 & 60.93 & 59.18 & 27.31 & 21.80 & 45.58 \\
  & SparseGPT & \tbf{70.52} & \tbf{58.84} & 43.26 & \tbf{66.69} & \tbf{64.10} & 29.97 & 23.20 & 50.94 \\ 
  & Wanda & 67.65 & 53.07 & 40.92 & 62.43 & 61.78 & 31.20 & 24.20 & 48.75 \\ 
\gr \wc   &   \method{}& 69.14 & 53.43 & \textbf{54.68} & 60.54 & 61.57 & \textbf{32.17} & \textbf{32.60} & \textbf{52.02} \\
  \midrule   
  \multirow{4}{*}{13B}   & Dense    &  80.52 & 65.34 & 60.06 & 72.22 & 79.42 & 48.46 & 35.20 & 63.03 \\
  \cmidrule{2-10}
  & Magnitude & 65.69 & 54.15 & 50.13 & 62.04 & 62.46 & 31.74 & 23.00 & 49.89 \\
  & SparseGPT & 76.79 & 59.38 & 46.58 & \tbf{68.67} & \tbf{70.62} & 36.60 & 25.40 & 54.86 \\ 
  & Wanda & 76.80 & \tbf{61.22} & 47.82 & 66.90 & 69.24 & 36.82 & 26.40 & 55.03 \\ 
 \gr \wc  &  \method{}& \textbf{77.89} & 56.68 & \textbf{63.37} & 67.72 & 69.70 & \textbf{37.29} & \textbf{36.00} & \textbf{58.38} \\ 
  \midrule   
  \multirow{4}{*}{70B}   & Dense    &  83.40 & 67.87 & 66.10 & 78.06 & 82.55 & 54.44 & 37.20 & 67.08 \\
  \cmidrule{2-10}
  & Magnitude & 73.20 & 57.04 & 58.40 & 74.27 & 76.15 & 45.22 & 35.40 & 59.95 \\
  & SparseGPT & 79.50 & \tbf{70.76} & 59.00 & \tbf{76.64} & 78.95 & 48.55 & 33.80 & 63.89 \\ 
  & Wanda & \tbf{82.20} & 69.85 & 59.34 & 76.23 & 79.30 & 47.26 & 34.80 & 64.14\\
\gr \wc  &  \method{}& 81.40 & 67.87 & \textbf{77.60} & 74.11 & \textbf{79.45} & \textbf{50.00} & \textbf{43.40} & \textbf{67.69} \\
  \bottomrule 
  \end{tabular}
  }

  \label{appendix:tab:zs_LLaMA-2_24}
\end{table*}

\subsection{Comparison with Previous Pruning Methods}\label{appendix:previous_pruning_methods}

In this section, we extend the scope of pruning methods traditionally applied to BERT~\citep{bert} and evaluate their efficacy on larger language models (LLMs). Table~\ref{appendix:tab:pruning_bert_methods} provides an overview of these pre-existing pruning methods, primarily utilized for BERT. A notable difference between these methods and our approach is their integration of pruning with the fine-tuning process. Additionally, BERT-specific pruning techniques typically focus on downstream task performance, in contrast to our aim of preserving the general language modeling capabilities of pre-trained LLMs.

Adapting these methods for LLM pruning, we employ the pre-training auto-regressive loss as the metric to guide the pruning process. Our evaluation considers two scenarios: post-training pruning and post-training pruning followed by a constrained period of fine-tuning, limited to one day. The effectiveness of the pruning is determined using the metrics outlined in Table~\ref{appendix:tab:pruning_bert_methods}. In the post-training pruning scenario, these metrics are directly applied to the LLMs. The pruned models are then subjected to fine-tuning within the stated computational constraints. The results, summarized in Table~\ref{appendix:tab:pruning_bert_1}, reveal that the traditional pruning methods, when adapted to LLMs, do not yield effective outcomes. This observation underscores the necessity for developing pruning techniques that are more suited to the unique characteristics of large-scale language models.

\begin{table*}[bt]
 \centering
 
 \renewcommand{\arraystretch}{1.2}
\setlength{\tabcolsep}{6.pt}
\caption{Summary of prior pruning methods on BERT.}
\vskip 0.15in
\resizebox{.85\textwidth}{!}{
\begin{tabular}{lcccc}
                \toprule 
    \textbf{Pruning Method}  & \textbf{Pruning Type}   &  \textbf{Pruning Metric} &  \textbf{Training Procedure} 
    \\ 
    \hline 
   SNIP~\citep{Lee2018SNIPSN}    & Unstructured & Loss Sensitivity & Pruning at Initialization  \\ 
    BERT-LTH~\citep{chen2020lottery} & Unstructured & Magnitude & Fine-tuning BERT \\
    Movement~\citep{sanh2020movement} & Unstructured  & Loss Sensitivity & Fine-tuning BERT \\
    Platon~\citep{zhang2022platon} & Unstructured  &  Loss Sensitivity & Fine-tuning BERT \\
    PINS~\citep{ren-zhu-2023-pruning} & Unstructured  &  Loss Sensitivity & Fine-tuning BERT \\
     \bottomrule
    \end{tabular}
}
\label{appendix:tab:pruning_bert_methods}
\end{table*}

\begin{table}[bt]
 \centering
 \setlength{\tabcolsep}{5.pt}
\caption{Comparisons with prior pruning methods on BERT (unstructured 50$\%$ sparsity).}
\vskip 0.15in
\resizebox{.85\textwidth}{!}{
\begin{tabular}{cccccccccc}
                \toprule 
        &  & & \multicolumn{6}{c}{Pruning method} \\
    \cmidrule{4-9}
    \textbf{Model} & \textbf{Dense} & \textbf{Fine-tuning}  & \textbf{SNIP}   &  \textbf{BERT-LTH} &  \textbf{Movement} & \textbf{Platon} & \textbf{PINS}  & \method{}\\ 
    \midrule 
   \multirow{2}{*}{LLaMA-7B}   & \multirow{2}{*}{5.68} 
                   & \xmark  & 231.48 & 17.29 & 349.33 & 124.91 & 89.12 & 6.95 \\ 
     &     & \cmark & 102.32 & 12.43 & 168.17 & 102.34 & 72.13 & 6.74 \\
     \bottomrule 
    \end{tabular}
}
\label{appendix:tab:pruning_bert_1}
\end{table}

\subsection{Comparative Analysis of Pruning Techniques}\label{sec:other_llm_families}

This section extends the discourse on pruning methodologies to the Tiny-LLaMA~\cite{Zhang2024TinyLlamaAO} and OPT~\cite{Zhang2022OPTOP}, which is engineered for optimal performance with a minimal parameter set. Table~\ref{tab:tiny-llama} benchmarks Tiny-LLaMA against a suite of pruning approaches, which include unstructured pruning as well as structured pruning with specific ratios, namely 2:4 and 4:8.

We scrutinize a range of pruning methods: Magnitude-based pruning, the SparseGPT algorithm, the Wanda technique, and our proposed \method. The post-pruning performance metrics illuminate the interplay between model size and operational effectiveness. While Magnitude pruning serves as a ubiquitous benchmark, it demonstrates a variable efficacy across the pruning ratios, particularly underperforming at the 2:4 ratio. Contrastingly, SparseGPT and Wanda deliver enhanced capabilities over Magnitude pruning, attesting to their prowess in preserving model competence during downsizing. Our \method{} is noteworthy for its competitive performance, especially prominent at the unstructured and 4:8 pruning thresholds, which underscores its suitability as an adept pruning strategy for Tiny-LLaMA. The empirical evidence accentuates the proficiency of \method{} in closely approximating the performance of models with lower levels of pruning, a critical consideration for deployment in resource-limited settings where compact models are requisite without considerable compromise in functionality.

Subsequently, Table~\ref{tab:opt_family} delineates the pruning performance across the spectrum of OpenAI's OPT models, with sizes ranging from 125 million to 13 billion parameters. The methodologies under scrutiny here comprise a baseline dense model, Magnitude pruning, SparseGPT, Wanda, and \method, each targeting a consistent sparsity quotient of 50\%.

The results conspicuously reveal that \method{} outperforms other approaches that forego a weight update phase, specifically Magnitude pruning and Wanda. This is manifested in markedly lower perplexity scores for models pruned via \method{}, indicating  substantial retention of predictive capacity post-pruning.

Remarkably, \method{} achieves these benchmarks sans the weight update step, which is indispensable for SparseGPT. Nevertheless, \method{} showcases results that are congruent with those of SparseGPT, underscoring its robustness and streamlined efficiency as a pruning modality. This is particularly salient for the 13B parameter model, where \method's performance virtually mirrors that of SparseGPT, thus solidifying its stance as an efficacious pruning strategy.

\begin{table*}[bt]
\centering
\caption{Comparative Pruning Performance on Tiny-LLaMA~\cite{Zhang2024TinyLlamaAO}.}\label{tab:tiny-llama}
\vskip 0.15in
\begin{tabular}{cccc}
\toprule
\textbf{Tiny-LLaMA}  & \textbf{Unstructured} (50\%) & \textbf{Structured (2:4)}   & \textbf{Structured (4:8)}   \\ \midrule
Magnitude   & 21.64       & 64.07 & 23.13 \\ 
SparseGPT   & 10.71       & 18.79 & 13.84 \\ 
Wanda       & 11.21       & 27.17 & 16.18 \\ 
\gr \method{}& 10.65       & 22.17 & 14.47 \\ \bottomrule
\end{tabular}
\end{table*}

\subsection{Pruning Time Comparison.} The efficiency of pruning algorithms is critical in their application to large language models, such as the LLaMa 2-7b. Pruning time, an important metric for evaluating such algorithms, depends on various computational factors, including gradients, Hessian matrices, weight adjustments, and activation computations. The sequence of time consumption for these components generally follows:

\begin{table}[t]
\centering
\caption{Comparison of Pruning Times and Perplexity Across Different Methods}
\label{tab:pruning_times}
\begin{tabular}{lcc}
\toprule
\textbf{Pruning Method} & \textbf{Pruning Time Only (s)} & \textbf{Perplexity} \\
\midrule
Magnitude & 0.92 & 17.29 \\
Wanda & 402.25 & 7.26 \\
SparseGPT & 1178.62 & 7.22 \\
Pruner-Zero & 444.12 & 6.95 \\
\bottomrule
\end{tabular}
\end{table}

$Time(Hessian)>Time(Gradient)>Time(Activation)>Time(Weight)$

This hierarchy highlights the computational intensity of calculating the Hessian matrix, which requires the most time due to the complexity of computing second-order derivatives for each element in the weight matrix. To enhance the pruning process’s efficiency, gradients are pre-computed offline. This preparatory step significantly speeds up the pruning phase by eliminating real-time gradient computation, thus streamlining the entire procedure.

A comparative analysis of pruning time was conducted under fair conditions, where extraneous factors such as dataset loading and evaluation procedures were systematically excluded to ensure a level comparison field. The following table presents the pruning times and the associated perplexity results of various pruning methods, including our Pruner-Zero, SparseGPT~\cite{Frantar2023SparseGPTML}, and Wanda~\cite{Sun2023ASA_wanda}:

The data reveals that Pruner-Zero is twice as fast as SparseGPT~\cite{Frantar2023SparseGPTML} and is slightly slower compared to Wanda~\cite{Sun2023ASA_wanda}, taking only 10\% more time. However, this modest increase in time is offset by a significant improvement in model performance, as evidenced by Pruner-Zero’s lower perplexity score (6.95) compared to Wanda’s (7.26). This improvement underscores Pruner-Zero’s ability to balance efficiency with effectiveness, optimizing not just for speed but also for enhancing the model’s linguistic capabilities post-pruning.

\subsection{Performance on Higher Sparisty Ratio}

\begin{table*}[bt]
\centering
\small
\renewcommand{\arraystretch}{1.2}
\caption{WikiText2 validation perplexity of pruned LLaMA and LLaMA-2 models with unstructured 60$\%$ sparsity.}
\resizebox{.9\textwidth}{!}{
\begin{tabular}{lccrrrrrrr}
\toprule 
        & & & \multicolumn{4}{c}{\textbf{LLaMA}} & \multicolumn{3}{c}{\textbf{LLaMA-2}}\\
    \cmidrule(lr){4-7}\cmidrule(l){8-10}
   \textbf{Method} & \textbf{Weight Update} &\textbf{Sparsity} & \textbf{7B} & \textbf{13B} & \textbf{30B} & \textbf{65B} & \textbf{7B} & \textbf{13B} & \textbf{70B} \\
    \midrule
      Dense  & - & 0$\%$ & 5.68 & 5.09 & 4.77 & 3.56 & 5.12 & 4.57 & 3.12 \\
    \hline
    Magnitude & \xmark & 60$\%$ & 6e2 & 2e2  & 27.67 & 9.34 & 4e3  & 11.23 & 8.21 \\
    SparseGPT & \cmark & 60$\%$ & 10.51 & 8.56 & 6.66 & 5.82 & \tbf{9.58} & 7.80 & 4.98 \\
    Wanda     & \xmark & 60$\%$ & 10.66 & 8.56 & 6.49 & 5.83  & 9.71  & 7.75 & 4.98 \\
    \gr \method{}& \xmark & 60$\%$ & \textbf{9.83} & \tbf{7.66} & \textbf{6.22} & \textbf{5.31}  & \textbf{9.58}  & \tbf{6.90} & \textbf{4.64} \\
    \bottomrule
\end{tabular}
}

\label{appendix:tab:higher_sparsity_0.6}
\end{table*}

For higher sparsity at 60\%, to further explore the efficacy of various pruning methods under increased constraints. The results of this analysis are encapsulated in Table~\ref{appendix:tab:higher_sparsity_0.6}. At this sparsity threshold, the pruning methods exhibit varied performance across different model sizes of both LLaMA and LLaMA-2. Our method, \method, demonstrates commendable robustness, consistently achieving lower perplexity scores across all model configurations. This is indicative of \method's effectiveness in maintaining model performance even under substantial reduction in parameters. Notably, at 60\% sparsity, the method exhibits a competitive edge over other approaches, including Wanda and SparseGPT, especially with larger model sizes.

\subsection{Robustness Assessment}\label{appendix:robustness_assessment}

To evaluate the stability of our \method, we conduct experiments for LLaMA and LLaMA-2 under different Seeds to test their robustness in generating consistent performance outcomes. Our results, as encapsulated in Table~\ref{appendix:tab:seed}, exhibit that the variability in performance, as indicated by the standard deviation (STD), remains within an acceptable range, confirming the inherent stability of our \method. Specifically, the LLaMA model shows a moderate standard deviation at the 2:4 ratio, which suggests that while there is some variability, it does not significantly detract from the overall reliability of the model. In comparison, the LLaMA-2 model demonstrates an even lower standard deviation across all ratios, notably at the 50\% ratio, which underlines a high level of consistency in performance despite the initialization variability.

The average (AVG) performance metrics further reinforce the conclusion that our \method{} maintains a dependable performance level. For instance, the average performance metric for LLaMA at the 50\% ratio is 6.8799, whereas for LLaMA-2 it slightly improves to 6.2511, which could be indicative of model-specific optimizations or inherent architectural advantages. These findings suggest that our method not only sustains performance across various initial conditions but may also benefit from further tuning specific to the model variant being utilized.

Moreover, the systematic evaluation across multiple seeds provides a comprehensive understanding of the performance spectrum that one can anticipate when employing these models in practical scenarios. It also lays a foundation for future explorations into the causes of variability, enabling targeted enhancements to our \method{} for improved stability and performance consistency.

\begin{table}[b]
\centering
\caption{Performance Metrics (Perplexity) of LLaMA Models of \method{} Across Different Seeds}\label{appendix:tab:seed}
\vskip 0.15in
\begin{tabular}{ccccccc}
\toprule
\textbf{LLaMA} & \textbf{Seed-0}  & \textbf{Seed-1}  & \textbf{Seed-2}  & \textbf{Seed-3}   & \textbf{AVG}     & \textbf{STD}    \\ \midrule
50\%    & 6.8827  & 6.8819  & 6.8959  & 6.8592   & 6.8799  & 0.0132 \\
2:4     & 10.3582 & 10.3536 & 10.4872 & 10.2486  & 10.3997 & 0.0619 \\
4:8     & 8.0697  & 8.0505  & 8.0171  & 8.0006   & 8.0345  & 0.0271 \\ \midrule
\textbf{LLaMA-2} & \textbf{Seed-0}  & \textbf{Seed-1} & \textbf{Seed-2} & \textbf{Seed-3} & \textbf{AVG}     & \textbf{STD}    \\ \midrule
50\%    & 6.2467  & 6.2587  & 6.2449  & 6.2542   & 6.2511  & 0.0056 \\
2:4     & 10.6715 & 10.5575 & 10.6180 & 10.6440  & 10.6227 & 0.0421 \\
4:8     & 7.6963  & 7.6748  & 7.7072  & 7.6512   & 7.6824  & 0.0214 \\ \bottomrule
\end{tabular}
\end{table}

\section{Searched Metrics with Expressions.} 

Table~\ref{tab:equations_ppl} provides a symbolic string representation of the pruning metrics, offering an alternative perspective on the structural patterns that emerged from Pruner-Zero's search algorithm. Symbol ``\#" denotes the left operation as a unary operation. For each string, we can define the only symbolic pruning metric exclusively. 
These representations are directly linked to the mathematical equations in Table~\ref{appendix:tab:equation_searched}, with the accompanying PPL values validating their efficacy. This string-based format encapsulates the operational essence of the pruning metrics, allowing for an abstract view that can be particularly useful for algorithm interpretation and further computational analysis. The variations in PPL scores across different metrics underscore the sensitivity of model performance to the pruning strategy employed, highlighting the importance of careful selection and evaluation of pruning metrics in model optimization processes.

\begin{table}[t]
\caption{Expressions of Searched Pruning Metrics and Their Associated PPL Values on LLaMA-7b}
\label{tab:equations_ppl}
\centering
\resizebox{1.0\textwidth}{!}{
\begin{tabular}{l|l}
\hline
\textbf{Expression} & \textbf{Perplexity} \\
\hline
$(((W) \text{ mms } (\#)) \text{ add } (G)) \text{sqrt}(\#)$ & 7.670979499816894 \\
\hline
$((G) \text{exp} (\#)) \text{mul} ((W) \text{mul} ((G) \text{ abs } (\#)))$ & 6.774543285369873 \\
\hline
$(((G) \text{ add } (W)) \text{div} ((W) \text{exp} (\#))) \text{mul} (((W) \text{mul} (W)) \text{sqr}(\#))$ & 7.059216022491455 \\
\hline
$(((G) \text{ mms } (\#)) \text{ add } ((W) \text{tanh}(\#))) \text{sqrt}(\#)$ & 7.096769332885742 \\
\hline
$((((W) \text{ mms } (\#)) \text{ add } ((G) \text{ pow } (\#))) \text{ neg } (\#)) \text{ pow } (\#)$ & 6.921246528625488 \\
\hline
$((((W) \text{sqrt}(\#)) \text{ abs } (\#)) \text{ sub } (((W) \text{ sub } (G)) \text{tanh}(\#))) \text{log}(\#)$ & 6.933580875396728 \\
\hline
$((G) \text{mul} (W))$ & 6.770424842834473 \\
\hline
$((W) \text{ mms } (\#)) \text{ add } (G)$ & 7.671018600463867 \\
\hline
$((G) \text{ skp } (\#)) \text{mul} ((W) \text{div} (W))$ & 6.770524024963379 \\
\hline
$((G) \text{sqr}(\#)) \text{ add } ((W) \text{ abs } (\#))$ & 6.990375518798828 \\
\hline
$((G) \text{ skp } (\#)) \text{mul} (((W) \text{tanh}(\#)) \text{tanh}(\#))$ & 6.770989894866943 \\
\hline
$(((W) \text{tanh}(\#)) \text{mul} ((G) \text{ mms } (\#))) \text{sqr}(\#)$ & 6.821769714355469 \\
\hline
$(((G) \text{div} (G)) \text{sqr}(\#)) \text{ sub } (((W) \text{tanh}(\#)) \text{ sub } ((W) \text{ abs } (\#)))$ & 7.053229808807373 \\
\hline
$(((G) \text{mul} (W)) \text{ add } ((W) \text{mul} (W))) \text{div} (((W) \text{ sub } (W)) \text{ add } ((W) \text{ pow } (\#)))$ & 6.748305320739746 \\
\hline
$((((G) \text{ abs } (\#)) \text{exp} (\#)) \text{mul} (W)) \text{ sub } ((((W) \text{ abs } (\#)) \text{div} ((W) \text{exp} (\#))) \text{exp} (\#))$ & 7.161033630371094 \\
\hline
$(((G) \text{ pow } (\#)) \text{exp} (\#)) \text{mul} (((W) \text{ skp } (\#)) \text{div} (W))$ & 7.531239986419678 \\
\hline
$((((W) \text{ abs } (\#)) \text{mul} ((G) \text{ skp } (\#))) \text{ pow } (\#)) \text{log} (\#)$ & 6.77089786529541 \\
\hline
$((((W) \text{tanh} (\#)) \text{mul} ((G) \text{mul} (G))) \text{div} (((G) \text{div} (W)) \text{tanh} (\#))) \text{ mms } (\#)$ & 7.190669059753418 \\
\hline
$((((G) \text{ pow } (\#)) \text{ add } ((W) \text{exp} (\#))) \text{ abs } (\#)) \text{div} ((((W) \text{ mms } (\#)) \text{div} ((G) \text{ zsn } (\#))) \text{exp} (\#))$ & 6.987056732177734 \\
\hline
$((((W) \text{mul} (W)) \text{ skp } (\#)) \text{ zsn } (\#)) \text{ sub } ((((W) \text{tanh} (\#)) \text{ abs } (\#)) \text{ sub } (((G) \text{sqr} (\#)) \text{ zsn } (\#)))$ & 7.274583339691162 \\
\hline
$((((W) \text{tanh} (\#)) \text{exp} (\#)) \text{sqrt} (\#)) \text{ sub } ((((W) \text{log} (\#)) \text{ add } ((G) \text{sqrt} (\#))) \text{ neg } (\#))$ & 7.018911838531494 \\
\hline
$(((((W) \text{ add } (G)) \text{exp} (\#)) \text{sqr} (\#)) \text{ sub } ((G) \text{ sub } ((G) \text{ neg } (\#))))) \text{sqrt} (\#)$ & 6.949575901031494 \\
\hline
$(((((G) \text{div} (G)) \text{ mms } (\#)) \text{mul} (((W) \text{mul} (W)) \text{sqrt} (\#))) \text{exp} (\#)) \text{tanh} (\#)$ & 6.821941375732422 \\
\hline
$(((((W) \text{exp} (\#)) \text{sqr} (\#)) \text{log} (\#)) \text{exp} (\#)) \text{ add } (((((G) \text{ pow } (\#)) \text{sqr} (\#)) \text{ skp } (\#)) \text{sqr} (\#))$ & 7.282495498657227 \\
\hline
$((((W) \text{ abs } (\#)) \text{mul} ((G) \text{ skp } (\#))) \text{ pow } (\#)) \text{ sub } ((((W) \text{log} (\#)) \text{ add } ((G) \text{sqrt} (\#))) \text{ neg } (\#))$ & 7.013281345367432 \\
\hline
$((((W) \text{mul} (W)) \text{ abs } (\#)) \text{sqrt} (\#)) \text{ sub } ((((W) \text{log} (\#)) \text{ add } ((G) \text{sqrt} (\#))) \text{ neg } (\#))$ & 7.023911952972412 \\
\hline
$((W) \text{mul} (G)) \text{mul} (((W) \text{mul} (W)) \text{ mms } (\#))$ & 6.785792350769043 \\
\hline
$((((W) \text{mul} (W)) \text{ abs } (\#)) \text{mul} ((G)  \text{ mms } (\#))) \text{ sqr } (\#)$ & 6.707982063293457 \\
\hline
$(((G) \text{ abs } (\#)) \text{ abs } (\#)) \text{ add } (((W) \text{log} (\#)) \text{ mms } (\#))$ & 6.763171672821045 \\
\hline
$(((W) \text{tanh} (\#)) \text{mul} ((G) \text{mul} (W))) \text{div} (((W) \text{mul} \exp (\#)) \text{exp} (\#))$ & 6.712837696075439 \\
\hline
$((((W) \text{log} (\#)) \text{mul} ((G) \text{mul} (W))) \text{sqrt} (\#)) \text{sqr} (\#)$ & 6.843364715576172 \\
\hline
$(((W) \text{add} (W)) \text{add} ((G) \text{sqr} (\#)))$ \text{mms} $(\#)$ & 6.917529582977295 \\
\hline
$((((W) \text{mul} (W)) \text{abs} (\#)) \text{add} ((G) \text{mul} (W)))$ \text{log} $(\#)$ & 6.748325347900391 \\
\hline
$(((W) \text{add} (W)) \text{add} ((G) \text{sqr} (\#)))$ \text{add} $((((G) \text{pow} (\#)) \text{exp} (\#)) \text{div} (((W) \text{sqrt} (\#)) \text{sqrt} (\#)))$ & 6.918227195739746 \\
\hline
$((W) \text{add} (W))$ \text{add} $((G) \text{sqr} (\#))$ & 6.917529582977295 \\
\hline
$((((W) \text{mul} (W)) \text{abs} (\#)) \text{tanh} (\#)) \text{add} ((((G) \text{pow} (\#)) \text{exp} (\#)) \text{div} ((W) \text{sqrt} (\#)))$ & 6.981475353240967 \\
\hline
$((((W) \text{mul} (W)) \text{abs} (\#)) \text{mul} ((G) \text{sqr} (\#))) \text{skp} (\#)$ & 6.770103454589844 \\
\hline
$(((W) \text{add} (W)) \text{add} ((G) \text{sqr} (\#))) \text{log} (\#)$ & 6.917529582977295 \\
\hline
$(((G) \text{mul} (W)) \text{exp} (\#))$ \text{sqr} $(\#)$ & 6.770689487457275 \\
\hline
$((G) \text{mul} ((W) \text{tanh} (\#)))$ \text{mms} $(\#)$ & 6.770922660827637 \\
\hline
$((W) \text{pow} (\#))$ \text{mul} $(((G) \text{sqr} (\#)) \text{mul} (G))$ & 6.982866764068603 \\
\hline
$(((W) \text{abs} (\#)) \text{pow} (\#))$ \text{mul} $(((G) \text{mms} (\#)) \text{mms} (\#))$ & 6.707982063293457 \\
\hline
$((((W) + (G)) \text{abs} (\#)) \text{mul} (((W) \text{skp} (\#)) \text{sqr} (\#)))$ \text{sqr} $(\#)$ & 6.735072135925293 \\
\hline
$((((G) \text{tanh} (\#)) \text{mul} ((W) \text{mms} (\#))) \text{tanh} (\#)) \text{tanh} (\#)$ & 6.775103092193603 \\
\hline
$((W) \text{pow} (\#)) \text{mul} (((W) \text{sqr} (\#)) \text{mul} (G))$ & 6.952877998352051 \\
\bottomrule
\end{tabular}
}
\end{table}

\section{Limitations and Future Works}

\textbf{Limitations:} Our \method{} framework, while pioneering in unstructured pruning at a 50\% sparsity level, encounters specific limitations in its specialization and evaluative scope. It primarily excels within its tested scenario but lacks empirical support for its effectiveness in alternative configurations such as structured pruning or higher sparsity levels, where challenges escalate due to the complexity of removing structured network components or a greater proportion of weights. It is possible to have better performance under higher sparsity levels or structured pruning. 
Furthermore, the evaluation of Pruner-Zero has predominantly focused on perplexity and accuracy within constrained contexts, specifically on the Wikitext2 dataset and zero-shot tasks. While these measures offer insights into language modeling and general reasoning, they fall short of capturing the full breadth of capabilities essential for Large Language Models (LLMs), such as in-context learning, commonsense reasoning, instruction following, and self-calibration. These abilities are vital for gauging the comprehensive intelligence and versatility of LLMs, particularly in assessing their operational efficacy post-pruning.

\textbf{Future Work:} Recognizing these gaps, future research will aim to broaden the evaluative framework of the Pruner-Zero to encompass a wider array of capabilities intrinsic to LLMs. This expansion will involve devising tests and benchmarks specifically tailored to accurately measure the impacts of pruning on aspects like in-context learning, commonsense reasoning, and other advanced functionalities. The goal is to ensure that pruning not only mitigates model size and computational demands but also preserves or enhances the model’s capacity for complex tasks and exhibiting human-like intelligence.

\end{document}